\documentclass[journal]{IEEEtran}

\ifCLASSINFOpdf

\else

\fi

\usepackage{graphicx}
\usepackage{comment}
\usepackage{amsmath,amssymb}
\usepackage{color}
\usepackage{hyperref}
\usepackage{times}
\usepackage{epsfig}
\usepackage{colortbl}
\usepackage{multirow}
\usepackage{array}
\usepackage{booktabs}
\usepackage{diagbox}
\usepackage{float}
\usepackage{epstopdf}
\usepackage[misc]{ifsym}

\newfloat{figtab}{htb}{fgtb}
\makeatletter
  \newcommand\figcaption{\def\@captype{figure}\caption}
  \newcommand\tabcaption{\def\@captype{table}\caption}
\makeatother

\begin{document}

\title{Clicking Matters:\\Towards Interactive Human Parsing}

\author{Yutong Gao*, %~\IEEEmembership{Student Member, IEEE}
        Liqian Liang*, %~\IEEEmembership{Student Member, IEEE}
        Congyan Lang, %~\IEEEmembership{Member, IEEE}
        Songhe Feng, %~\IEEEmembership{Member, IEEE}
 %       Tao Wang,%~\IEEEmembership{Student Member, IEEE}
        Yidong Li,
        Yunchao Wei %~\IEEEmembership{Member, IEEE}
        % <-this % stops a space
\thanks{Yutong Gao*, Liqian Liang*, Congyan Lang \emph{(corresponding author)}, Songhe Feng and Yidong Li are with the School of Computer and Information Technology,
 Beijing Jiaotong University, Beijing 100044, China. E-mail:(
ytgao92@bjtu.edu.cn; lqliang@bjtu.edu.cn; cylang@bjtu.edu.cn; shfeng@bjtu.edu.cn; ydli@bjtu.edu.cn).
Yunchao Wei is with the University of Technology Sydney, Sydney, NSW, Australia. E-mail:(wychao1987@gmail.com).  ``*'' indicates both authors contributed equally to this research. }}

\markboth{Journal of IEEE Transactions On Multimedia}%,~Vol.~14, No.~8, December~2020
{Shell \MakeLowercase{\textit{et al.}}: Towards Interactive Human Parsing}

\maketitle

\begin{abstract}
In this work, we focus on Interactive Human Parsing (IHP), which aims to segment a human image into multiple human body parts with guidance from users' interactions. This new task inherits the class-aware property of human parsing, which cannot be well solved by traditional interactive image segmentation approaches that are generally class-agnostic. To tackle this new task, we first exploit user clicks to identify different human parts in the given image. These clicks are subsequently transformed into semantic-aware localization maps, which are concatenated with the RGB image to form the input of the segmentation network and generate the initial parsing result. To enable the network  to better perceive user's purpose during the correction process, we investigate several principal ways for the refinement, and reveal that random-sampling-based click augmentation is the best way for promoting the correction effectiveness. Furthermore, we also propose a semantic-perceiving loss (SP-loss) to augment the training, which can effectively exploit the semantic relationships of clicks for better optimization. To the best knowledge, this work is the first attempt to tackle the human parsing task under the interactive setting. Our IHP solution achieves 85\% mIoU on the benchmark LIP, 80\% mIoU on PASCAL-Person-Part and CIHP, 75\% mIoU on Helen with only 1.95, 3.02, 2.84 and 1.09 clicks per class respectively. These results demonstrate that we can simply acquire high-quality human parsing masks with only a few human effort. We hope this work can motivate more researchers to develop data-efficient solutions  to IHP in the future.
\end{abstract}


\IEEEpeerreviewmaketitle

\section{Introduction}

\IEEEPARstart{H}{uman} parsing aims to assign each pixel with one of the pre-defined semantic categories of human body parts like arm, hair, shirt, \emph{etc}.
It is applied to various applications such as intelligent surveillance \cite{liang2015fashion}, human clothing retrieval \cite{liang2016clothes} and human-robot interaction~\cite{2016A,fang2019graspnet}.
Recently, deep learning based parsing methods have achieved much progress by exploring local context among adjacent human parts \cite{zhang2020part,FCM,Yang2018Parsing,Yuan2018OCNet,ce2p}
or global topological structure buried in the entire human body~\cite{li2020self,wang2020hierarchical,graphonomy}.
However, these deep methods are often hungry for large scale well annotated training data, which are costly to attain.
Even the state-of-the-art parsing accuracy currently is less than 60\% mIoU \cite{wang2020hierarchical},
and the parsing results tend to show low quality with much error like overlapping or coarse boundaries.
Towards the goal of applying human parsing to real world scenarios where large pose and appearance variation, occlusion, and bad illumination are quite prevailing, we consider further improving the parsing performance by introducing human interaction information as supervisions.

By exploiting human interaction for facilitating segmentation, Interactive Object Segmentation (IOS) has been widely studied.
In this task, the explored interaction manners mainly include clicks \cite{border1,extremecut,border2,border3,first,region,inside1}, scribbles~\cite{full2} and bounding boxes~\cite{grabcut}.
Though effective, the IOS methods aim at segmenting an independent object or object part specified by the user, which essentially perform binary classification of foreground vs. background, rather than aiming at recognizing what class the object or object part belongs to.
For example, the work \cite{extremecut} employs extreme points to locate an object of interest, and then casts the encoded inputs with point information and RGB information into a fully convolutional network to generate a foreground-background binary mask associated with the selected object.
Hence, IOS tackles a class-agnostic setting and ignores the semantic relationships among object parts, while human parsing pursues class-aware understanding of human body in nature.

\begin{figure}[t]
\setlength{\abovecaptionskip}{-0.2cm}
\setlength{\belowcaptionskip}{-0.2cm}
\begin{center}
   \includegraphics[width= 8.3cm,height = 3.4cm]{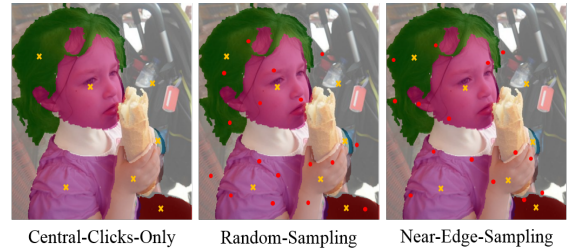}
\end{center}
\vspace{-1mm}
   \caption{
   \small{Illustration of three clicking simulation strategies.
   Yellow symbols ``$\times$'' indicate central clicks for each human part and background,
   while red dots indicate extra clicks.}
}
\label{fig:clickingsample}
\vspace{-3mm}
\end{figure}

In this work, we formulate a new Interactive Human Parsing (IHP) task, which aims to decompose a human to multiple body parts simultaneously with the guidance of a few user clicks.
To our best knowledge, this work is the first attempt to resolve human parsing under a human-machine interactive setting.
In contrast with IOS, the IHP task is featured by the following characteristics.
First, IHP tackles rich semantic categories. In other words, IHP is a class-aware task compared with IOS that is class-agnostic.
Secondly, IHP pursues densely distributed human body parts, \emph{i.e.}, classifying all the pixels into multiple human body parts simultaneously, while IOS only segments an object or object parts specified via user-machine interaction.
Thirdly, IHP is able to benefit from the interactive information between different human body parts.
That is, IHP can employ the semantic relationships of interactive clicks to learn a more discriminative feature embedding.

Similar to IOS, our IHP model also follows a pipeline of 1) encoding the user clicks to generate localization maps, and 2) concatenating RGB image with the localization maps to form the input of segmentation network. Therefore, an expected IHP model should also meet two criteria. First, with human interactions, the model can accurately produce high-quality parsing results. Second, when error appears, the model should allow iterative correction until result is satisfactory. Therefore, the main challenge of IHP lies in how to make the model perceive human's purpose, reducing human effort in clicking, the less the better.

To tackle IHP, in this work, we investigate several interaction simulation strategies during the training stage to find out the best way of making the learned model understand human's purpose during testing. In particular, we explore three clicking strategies, named \emph{Central-Clicks-Only}, \emph{Random-Sampling} and \emph{Near-Edge-Sampling}, respectively, as shown visually in Fig.~\ref{fig:clickingsample}.
The \emph{Central-Clicks-Only} targets at enabling the model to learn to locate each human part efficiently, where we follow prior works~\cite{first,superpixel,first-attention} to simply position simulated clicks around the central area of each human part.
The \emph{Random-Sampling} aims to make the model update the parsing result effectively based on human parts localization, where we randomly sample extra simulation clicks after making central simulation clicks.
Besides, we empirically find most of the errors in the initial parsing are located near human part edges. So, we further adopt a \emph{Near-Edge-Sampling} strategy by adding simulated clicks around human parts boundaries, aiming to make the model be robust to the near-edge regions during the refinement.
Through experiments, we interestingly find that the second strategy, Random-Sampling, is the best choice for simulating human clicks.
The detailed explanations of each strategy will be given in Section~\ref{strategies} and analysis experiments will be reported in Section~\ref{ablation}, respectively.

To further lift IHP performance, we additionally propose a semantic-perceiving loss (SP-loss) to better exploit the semantic relationships between human parts and expand the effect of the interactive clicks.
Inspired by the triplet loss \cite{triplet}, our SP-loss encourages the feature representation of each human part to be consistent with interactive information.
Interactive clicks are optimized to share their feature representations more widely within the same human part, while increasing the discrepancy of features in the adjacent parts.
With this loss, our model can interact with user clicks more efficiently, and is able to achieve good results with fewer clicks.

The major contributions of this paper are summarized in threefold.
\begin{enumerate}
\item We formulate a new problem, \emph{i.e.}, IHP, to study human parsing under the human-machine interactive setting to further boost parsing performance.
To our best knowledge, our work is the first of its kind.
\item We offer a simple but effective solution to the new problem by training an end-to-end deep IHP network.
Particularly, we explore three strategies to simulate user clicking behavior, and also propose a SP-loss to better exploit semantic relationships of interactive clicks among multiple human parts.
\item We conduct extensive experiments and well prove the effectiveness of our IHP model.
It achieves promising results on human parsing and face parsing datasets.
By user study, our method can achieve 85\% mIoU on LIP dataset~\cite{Liang2018Look} with 1.86 interactive clicks on average per human body part.
\end{enumerate}

\section{Related Work}

\subsection{Interactive Object Segmentation}
IOS aims to utilize user interaction to infer the region of interest in an image.
Compared with semantic segmentation, IOS is essentially a binary classification problem,
since it only segments the foreground object from the background under a class-agnostic setting.
This task has been explored for decades as it can complement fully-automated segmentation with user interaction.
Existing IOS methods can be roughly divided into three categories according to their adopted user interaction manner: bounding-box-based methods \cite{grabcut}, contour-based methods \cite{contour1,contour2}, and click-based methods\cite{first,extremecut,superpixel,first-attention,Nvidia-Object}.
Here we focus on the third group as they are the most popular.
%We apply clicks as supervision in our method to further boost human parsing performance.

Early click-based (or stroke-based) methods popularly adopt graph-based algorithms, including normalized cuts \cite{normalizedcut}, graph cut \cite{graphcut1,graphcut2,graphcut3}, geodesics \cite{geodesics1,geodesics2}, combinations of graph cut and geodesics~\cite{comb1,comb2}, and random walks \cite{randomwalk} for segmentation.
Recently, deep learning networks have achieved great success in click-based IOS,
starting from \cite{first} where user clicks are converted to Euclidean distance transformed maps and concatenated with color channels to be fed into FCN \cite{FCN} for fine-tuning.
Subsequent works strive to improve their user interaction transformation \cite{gaussian1,superpixel,Nvidia-Object,first-attention,full1},
use better training procedures \cite{gaussian2,region,BRS,r-BRS},
or enhance their interaction (clicking) simulation strategies \cite{border1,extremecut,border2,border3,inside1,in-out-side}.

The prior works like \cite{gaussian1,superpixel,Nvidia-Object} improve ways of transforming user interaction.
In particular, they encode clicks as Gaussian that is more conducive to extreme-click-based methods and encode all the clicks in a single channel.
\cite{superpixel} utilizes superpixels to generate the guidance map; \cite{Nvidia-Object} is based on \cite{extremecut}, and it uses dynamic edge dragging to strengthen the model's ability to correct error.
Some works focus on improving training procedures.
For example, \cite{region} proposes RIS-Net that expands the field-of-view of the input to capture local regional information in its neighborhood for local refinement;
\cite{BRS} and \cite{r-BRS} develop backpropagating refinement schemes that correct mislabeled pixels in the initial segmentation.
In terms of adopting better simulation clicking strategies, current methods explore two manners: placing clicks along the object edges \cite{border1,extremecut,border2,border3,Nvidia-Object}, or placing clicks inside or outside object regions \cite{first,region,inside1,superpixel,first-attention,in-out-side}.
The former clicking manner merely offers the position information for the model to identify explicit boundaries to segment the object of interest, while the latter provides both the binary labels (foreground or background) and the position information as reference for learning the range of user clicks to guide the model to locate the object.

Existing IOS methods are class-agnostic and perform object-level or instance-level segmentation, rather than class-aware semantic segmentation.
Directly applying them to object-part-level semantic segmentation, as is the case in human parsing,  would lead to sharp performance decrease due to redundancy of interactive information and confusion when assigning categories to multiple object parts.
To our best knowledge, our work is the first attempt to explore human parsing under the human-machine interactive setting.
Furthermore, we also explore various simulation clicking strategies to achieve good performance, and try to exploit semantic relationships of interactive clicks among multiple human parts.

\subsection{Human Parsing}
Traditional human parsing is a sub-problem of standard semantic segmentation, which requires pixel-level classification with body part labels rather than binary class-agnostic labels.
The current state-of-the-art is less than 60\% mIoU \cite{wang2020hierarchical} and the parsing quality is often far from satisfactory, with much error like overlapping or coarse boundaries.
Existing human parsing methods are mostly based on powerful semantic segmentation networks such as Deeplab V3+ \cite{deeplab}
which uses the ASPP module to expand the range of the receptive field of a CNN model, or PSP \cite{PSP} which adopts multi-level average pooling to aggregate long-range global information and local information into a single output feature.
In this paper, we build our framework on top of the widely-used Deeplab V3+ but with user interaction incorporated.

In \cite{Liang2015Human}, a Co-CNN architecture is developed to integrate cross-layer context, global image-level context, within-super-pixel context and cross-super-pixel neighborhood context;
a more powerful JPPNet~\cite{Liang2018Look}  jointly performs human parsing and pose prediction with high quality by exploiting multi-scale feature connections.
Some subsequent human parsing works explore local context among adjacent human parts \cite{FCM,Yang2018Parsing,Yuan2018OCNet,ce2p,zhang2020part} to achieve improved performance.
For example, \cite{Yang2018Parsing} proposes a new Parsing R-CNN which is able to represent the details of human instances with a region-based approach;
Ocnet \cite{Yuan2018OCNet} introduces an object context pooling (OCP) scheme, which represents each pixel by exploiting the set of pixels that belong to the same object category to enhance the relationship among them; \cite{ce2p} analyzes properties that benefit human parsing and integrate them into a CE2P framework; \cite{zhang2020part} develops a part-aware context network to generate adaptive contextual features for the various sizes and shapes of human parts.
In addition, some works strive to capture the global topologic structure buried in the entire human body \cite{graphonomy,li2020self,wang2020hierarchical} to boost parsing performance.
For exploiting the global topologic structure, \cite{graphonomy} proposes a new universal human parsing agent called Graphonomy, which incorporates hierarchical graph transfer learning upon the conventional parsing networks to predict all labels in one system without piling up the complexity;
\cite{wang2020hierarchical} proposes a hierarchical human parser which exploits the representational capacity of deep graph networks and also the hierarchical human structures.

In this work, we consider human parsing under the human-machine interactive setting and also propose a simple but effective solution where user clicks are smartly used to guide the model to achieve better parsing results.

\begin{figure*}[t]
\begin{center}
   \includegraphics[width=17.8cm,height = 3.5cm]{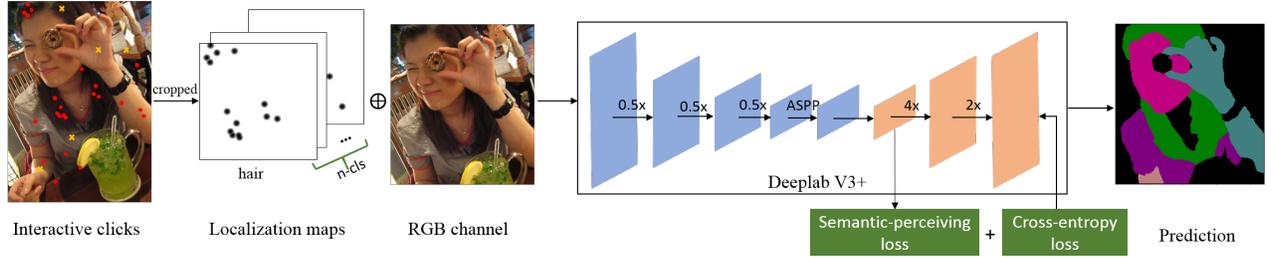}
\end{center}
\vspace{-5mm}
   \caption{
   Pipeline of our proposed framework for IHP.
   ``n-cls'' indicates the number of semantic classes.
   }
\label{fig:pipe}
\vspace{-3mm}
\end{figure*}

\section{Proposed Method}
In this section, we elaborate the proposed deep IHP framework.
We first give an overview of the framework.
Then, we describe the procedure of encoding interactive clicks, and also propose a SP-loss to assist learning explicit relationships among interactive clicks.
Finally, we explore three interaction simulation strategies to find a way with both effectiveness and efficiency.

\vspace{-3mm}
\subsection{Framework Overview}

Fig.~\ref{fig:pipe} shows the overall pipeline of our proposed framework for our new IHP task.
Our framework takes the RGB image together with a set of user-provided clicks as input and predicts a fine-grained human parsing mask.
The interactive clicks are first encoded into a tensor consisting of several localization maps, which is then concatenated with RGB channels of the original image to be fed into the segmentation network.
For the first layer of the network, we augment the number of channels of the convolutional kernel to fit input size.
As illustrated in Fig.~\ref{fig:pipe}, the overall network is trained end-to-end with the cross-entropy loss and our proposed SP-loss on four benchmarks,
LIP \cite{Liang2018Look}, CIHP \cite{pgn}, PASCAL-Person-Part \cite{pascalpart} and Helen \cite{face}.
For training, we simulate user clicks following an area-based clicking mode, \emph{i.e.}, to locate user clicks within objects or object parts rather than an edge-based one, namely along their edges.
This is because edge-based clicks as adopted in DEXTR \cite{extremecut} are hard to be assigned with explicit human body part labels due to dense category distribution, while using an area-based clicking mode can fully harvest the interactive information among multiple categories.

\vspace{-3mm}
\subsection{Interactive Click Encoding}

In the IHP problem, there should be no concept known as positive clicks and negative clicks like other IOS methods, due to its class-awareness.
Instead, all the interactive clicks have semantic labels, which can be used to classify among different human parts.
Suppose $C$ classes of human parts (including background) are pre-defined.
We denote the collection of interactive clicks by $U = \{U^i\}, i=0,1,\dots,C$,
where $U^i$ represents the click set for the $i_{th}$ class.
First, we transform the interactive clicks into a set of localization maps $\{T^i\}, i=0,1,\dots,C$, where each $T^i$ has the same width and height of the input images.
%where $T^i \in R^{w \times h}$, with $w$ and $h$ representing width and height of the input images.
For each pixel $p$ in $T^i$, the encoded value is defined as follows:
\begin{equation}
\small{ T^i[p] = \min_{q \in U^i} \|p-q\|_2^2, i = 0,1,\dots,C.}
\end{equation}

We construct localization maps from the user clicks based on \cite{first}, but
instead of using the 2-norm as in \cite{first}, we use its square form and further truncate the encoded value at 255 to adapt to the dense spatial distribution of human parts.

\subsection{Loss Function}

\noindent \textbf{Semantic-Perceiving Loss.}
A standard semantic segmentation model is trained only with the cross-entropy loss,
which is pixel-independent and lacks explicit constraints on the relationships among the pixels.
We consider exploiting the semantics in the interactive clicks to remedy the above shortcoming.
Hence, we propose a SP-loss, aiming to pull normal pixels closer to interactive clicks in the feature space
while advancing body-context-sensitiveness of the model to the semantics of the interactive clicks.

The proposed SP-loss is inspired by the triplet loss \cite{triplet}.
Assume every two adjacent human parts constitute an adjacent pair, and the corresponding feature set for each of the paired human parts is expressed as $I^m$ and $I^n$, where $m$ and $n$ are their semantic labels respectively.
We randomly sample a point $r$ in the part $m$, and denote the corresponding feature as a general feature $I_r^m$.
Then we adopt the average value of the features of all the interactive clicks in the same class $m$ as the positive feature $I_{avg}^m$.
Similarly, we attain the negative feature $I_{avg}^n$ for the class $n$.
Then we calculate the cosine similarities between two pairs, \emph{i.e.}, the general feature/positive feature and the general feature/negative feature,
which are $\frac{I_r^m \cdot I_{avg}^m}{||I_r^m||\cdot||I_{avg}^m||}$ and $\frac{I_r^m \cdot I_{avg}^n}{||I_r^m||\cdot||I_{avg}^n||}$,
denoted as $S_p$ and $S_n$ respectively.
For $C$ semantic human parts, the per-sample SP-loss is defined as
     \begin{equation}
\begin{split}
\setlength{\abovedisplayskip}{5pt}
\setlength{\belowdisplayskip}{0pt}
\small{l_{semantic} = \sum_{i=1}^{C}\sum_{j=1}^{n_i}\min\{log(S_{n_{ij}})-log(S_{p_{ij}})+d_g,0\},}
\end{split}
\end{equation}
where $i$ represents the human part classes, $n_i$ represents the corresponding adjacent areas for each human part $i$
and the first term in $max(,)$ represents that $log(S_{n_{ij}})$ should exceed $log(S_{p_{ij}})$ by at least $d_g$ margin.

\noindent \textbf{Total Loss.}
Besides the proposed SP-loss, we also include the balanced cross-entropy loss which has been shown well-performing in boundary detection \cite{balancedloss1,balancedloss2} into our total loss.
To better utilize the interactive information, we make a simple modification on this balanced loss.
%\textcolor{red}{Specifically, compared to a pixel frequency weighted loss, we add more weight for each semantic class according to the frequency of the interactive clicks that belong to this class.The pixel frequency loss term and the interactive point frequency loss term of our balanced cross-entropy loss are assigned with an equal weight of 0.5.}
Different from normal balanced cross-entropy loss that is only employs the pixel frequency weighted loss, we also stress the value of  interactive clicks frequency for each semantic class. Therefore, our modified balanced cross-entropy loss include a pixel frequency weighted loss term and an interactive click frequency loss term, both with an equal weight of 0.5.

Finally, the total loss of the proposed framework $L$ is the weighted sum of the
SP-loss term $L_{sp}$, \emph{i.e.}, average of $l_{semantic}$ for all training samples,
and the balanced cross-entropy loss term $L_{ce}$.
It is formulated as follows:
\begin{equation}
\begin{split}
\small{L =  L_{ce} + \lambda L_{sp},}
\end{split}
\end{equation}
where $\lambda$ is the hyper-parameter to balance the contribution of two loss terms.

\vspace{-3mm}
\subsection{Simulating User Interaction}
\label{strategies}

We consider three clicking simulation strategies for training, \emph{i.e.}, Central-Clicks-Only, Random-Sampling and Near-Edge-Sampling.
%\textcolor{red}{to investigate how to add prior knowledge while fixing the inference strategy would yield good effectiveness and efficiency.}
We define a set of mIoU values as \textbf{parsing standards} on LIP, CIHP,  PASCAL-Person-Part and Helen datasets (see Section~\ref{Evaluation_method}).
In evaluation, the interaction cost (average click number for each parsing category) to achieve the \textbf{parsing standards} reflects the model's interaction efficiency.

Before describing the specific clicking manner in each strategy, we first explain the concept of ``central clicks''.
In our method,
%\textcolor{red}{a click is added to each human body part to provide location information to help generate a parsing result.This type of click is placed in the central area of each human part which we call a ``central click''.}
a click placed in the central area of each human part is used for providing location information to help generate a parsing result. And we call this type of clicks ``central clicks''.
Formally, let $\mathcal{O} = \{O^i\}$ be the set of ground truth parts,
where $i = 0,1,\dots,C$ and $O^i$ denotes a set of connected components that belong to the $i_{th}$ class.
Each component $O^i_j$ in $O^i$ is a closed region, and any two points in $O^i_j$ can be connected by a broken line that belongs entirely to $O^i_j$.
Since there usually exists occlusion on the human body, we simulate the central area clicks at the component-level rather than part-level.
%Hence, we take each $O^i_j$ as a central click.
Hence, we place one central click corresponding to each $O^i_j$.
Then we define ``extra clicks'' as simulated clicks in addition to central clicks during training.
With the same operation of adding central clicks, the difference of the three strategies mainly lies in the way of adding extra clicks.

\noindent\textbf{Central-Clicks-Only.}
In this strategy, we simulate a central click for each component $O^i_j$ and do nothing else.
When selecting central clicks for each human part component, we first randomly sample a subset $S^i_j$ of each $O^i_j$, where $S^i_j$ contains $N_c$ points.
Then we pick a click from $S^i_j$, whose minimum distance from the corresponding boundary set achieves the maximum value, as the central click of $O^i_j$.
The set of central clicks $\mathcal{C}$ is formulated as
\begin{equation}
\begin{split}
\begin{aligned}
 \small{\mathcal{C} = \{c_i | c_i = \arg \max_{p} \min\|p-q\|_2, \forall p \in N^i_c, \forall q \in B^i_j\}.}
\end{aligned}
\end{split}
\label{form:central_only}
\end{equation}
Eqn.~(\ref{form:central_only}) suits human part components.
For the background area, we only select one click with $d_{margin}$ pixels away from the human part boundaries in an image.
Note that the central clicks defined in our paper do not indicate the real center points in mathematics, but can cover the variance while simulating user clicking patterns.

This strategy can reduce the model's dependence on interaction effort since no extra clicks are involved and users only need provide a central click to each body component.
But the resultant model cannot update the parsing result during refinement as it is trained only over monotonous central-type simulations.

\noindent\textbf{Random-Sampling.}
On the basis of Central-Clicks-Only, this  random-sampling strategy is designed to add extra clicks in a relatively casual manner.
Assume $N_r$ clicks are randomly sampled across the whole image.
The number of clicks for background is limited within $[N_{b1}, N_{b2}]$, since it often occupies a rather large proportion in an image and too many pixels sampled in background may overwhelm other pixels with specific semantics, leading to a model lacking discrimination capacity for recognizing multiple human parts.

This strategy simulates extra clicks at various contexts, aiming to benefit the model on interaction efficiency when dealing with diverse user clicks.
It is experimentally (see Section~\ref{ablation}) proved optimal with our standard.

\noindent\textbf{Near-Edge-Sampling.}
On the basis of Central-Clicks-Only, we also explore adding extra clicks around the edges.
We define a set of near-edge points $\mathcal{E}$, corresponding to pixels inside a circle with center being an edge point belonging to $B^i_j$ and radius of $d_{margin}$.
Then we randomly sample near-edge points from $\mathcal{E}$:
 \begin{equation}
\begin{split}
\begin{aligned}
\small{\mathcal{E} =\{p| \|p-q\|_2 < d_{margin}, \forall q \in B^i_j\}.}
\end{aligned}
\end{split}
\end{equation}
The number of clicks for background is also limited within $[N_{b1}, N_{b2}]$ for the same reason mentioned.

In this strategy, simulation clicks are sampled near edges of human parts where are error-prone, with the goal of assisting the model in refining the near-edge area.
The model trained with this strategy can make better refinement near human parts boundaries which are difficult for human users to annotate.

Examples of three clicking strategies are illustrated in Fig.~\ref{fig:clickingsample}.
We find that our model can generate fairly promising segmentation masks by only using the first strategy, as central clicks can provide semantic information and corresponding locations simultaneously.
However, as the number of interaction points increases,
the accuracy of the first strategy is no longer improved.
That is where Random-Sampling and Near-Edge-Sampling work and both of them can promote mIoU by a large margin.
The detailed analysis among three clicking strategies will be described in Section~\ref{ablation}.

\section{Experiments}

\subsection{Datasets \& Evaluation}
\label{Evaluation_method}
\noindent \textbf{Datasets.}
The \textbf{LIP} dataset \cite{Liang2018Look} is a large-scale benchmark for single human parsing, including 50,000 images with pixel-wise annotations on 19 semantic human part labels.
The images are cropped person instances from MS COCO dataset \cite{coco},
which are captured from real-world scenarios.
We use 30,462 images for training, 10,000 images for validation and 10,000 images for testing.

The \textbf{PASCAL-Person-Part} dataset  \cite{pascalpart} is a set of additional annotations for PASCAL-VOC-2010.
It goes beyond the original PASCAL object detection task by providing pixel-wise labels for six human body parts.
There are 3,535 annotated images, split into a separate training set of 1,717 images and a validation set of 1,818 images.

The \textbf{CIHP} dataset \cite{pgn} is a large-scale multi-human dataset with 38,280 diverse multi-human images.
It is partitioned into training, validation and test set, with 28,280, 5,000, 5,000 images, respectively.
Each image is labeled pixel-wise with 19 semantic human part categories and also instance-level identification.

The \textbf{Helen} dataset \cite{helen} is composed of 2,330 face images with densely-sampled, manually-annotated contours around eyes, eyebrows, nose, outer lips, inner lips and jawline, which is originally designed for landmark detection.
\cite{face} provides resized and roughly aligned pixel-level ground truth with 10 face part labels to benchmark the face parsing problem.
It is separated into three parts: 2,000 images for training, 230 for validation and 100 for testing.

\noindent\textbf{Evaluation Metrics.}
Human parsing is usually tested with mIoU between ground truth and the predicted segmentation mask. We also use F-measure statistic to evaluate the parsing result around human part boundaries.
However, IOS is differently evaluated since the segmentation results can be improved by users adding interactive clicks.

During testing, our interactive system first gives a prediction after central clicks are assigned to each semantic connected component except for background (initialization stage);
%, We name these clicks ``initial clicks''
then one correction click (with correct label and coordinates) is required around the central area of the largest wrongly labeled region in each round of correction (refinement stage).
%we name these clicks ``correction clicks''
We automatically generate central clicks/correction clicks by selecting the point with the largest margin from the component/wrongly-labeled-region boundaries among five random candidates to cover the variance of user patterns.
We also conduct user study by replacing these simulated clicks with real user inputs.

We set our \textbf{parsing standards} to 85\%, 80\%, 80\%, and 75\% on LIP, CIHP, PASCAL-Person-Part, and Helen datasets, respectively.
%The system judges whether the parsing result is good enough, meaning whether further refinement is required, by checking weather the mIoU of the current parsing reaches 85\%, 80\%, 80\%, and 75\% on LIP, CIHP, Pascal-Personal-part, and Helen, respectively.
Note that unlike when interacting with actual users, \emph{i.e.}, user study, during testing, our system will not stop adding correction clicks if the performance for any single image meet the \textbf{parsing standards}.

For the whole test set, one way of testing our interactive system is from image-level:
the average number of correction clicks to achieve the \textbf{parsing standards}, called ``Additional number of clicks'' (``Add.'' for short) hereinafter,
namely  $\frac{Additional\, number\, of\, clicks}{Total \,of \, test \,images}$;
the other way is from category-level:
the ``Average number of clicks per class'' (``Avg.'' for short) to achieve the \textbf{parsing standards},
namely $\frac{Total \,of \, clicks}{\sum_{i=1}^{N_t} \#\,classes\,occurrences}$, where $N_t$ indicates the number of test images and ``\#'' means ``per image''.

The ``Add.'' and ``Avg.'' of the model to achieve \textbf{parsing standards} reflect model's interaction efficiency.
Note it is not easy for existing IOS methods to reach the \textbf{parsing standards} when their ``Avg.'' is approximately the same as ours.

\subsection{Implementation Details}

We adopt the basic structure and network settings provided by Deeplab V3+ \cite{deeplab} and fine-tune the network pre-trained on ImageNet \cite{imagenet} and MS COCO \cite{coco}.
To adapt to our IHP task, we make the following modifications.
1) To simulate the user interactions, we assign one central click for background and each connected component of human part.
2) The user clicks are encoded as localization maps for $C+1$ classes.
After concatenating them with the original RGB image, the number of channels of the first convolution layer is changed from 3 to $C+4$.
The kennel size of the first convolution layer is set to $7\times7$.
3) The entire network is trained end-to-end on four benchmarks.
Especially, we select the feature maps with a stride of 4 in the decoder network to calculate the SP-loss.

During training, we adopt similar data augmentation with CE2P~\cite{ce2p}, \emph{i.e.}, random scaling (from 0.5 to 1.5), cropping and left-right flipping.
The initial learning rate is 7e$^{-4}$.
We adopt SGD optimizer with the momentum of 0.9 and weight decay of $5\times10^{-4}$.
To stabilize the inference, the resolution of the input remains identical as the original.
Our method is implemented by extending the Pytorch framework \cite{pytorch}.
All networks are trained on four TITAN XP GPUs.
Due to the GPU memory limitation, the batch size is set to 12.
We train the models for 300,000, 120,000 and 120,000 iterations on LIP, PASCAL-Person-Part and Helen dataset for good convergence, and the training time for each dataset is 5d, 20h and 20h, which are similar as many other IOS methods.
We set the hyper parameters $w$, $h$, $d_g$, $N_c$, $d_{margin}$, $N_{b1}$ and $N_{b2}$ to be 473, 473, 1.5,  5, 10, 3, 6 respectively.
Particularly, $d_{margin}$ is set to 3 on Helen since the face parts are small and sparse.
We try different values for $\lambda$, and finally set it to 3 for Near-Edge-Sampling and 1 for Random-Sampling on all datasets.

\subsection{Ablation Study}
\label{ablation}
%We conduct ablation experiments to explore the best localization and simulation number of central clicks and extra clicks during training on LIP, Pascal-Person-Part and Helen datasets respectively.
%1) For central clicks, we test two variants, \emph{i.e.}, central clicks and random clicks, to find an effective way to locate human parts.
%2) For extra clicks, we combine different clicking strategies with different extra clicks number to form diverse variants of our method in order to find an effective way to deal with diverse user clicks exist in refinement stage. Specifically, for Random-Sampling and Near-Edge-Sampling strategy, we conduct three variants trained with $1\sim15$, $1\sim45$ and optional $1\sim75$ extra clicks per-image separately. Plus the model trained with Central-Clicks-Only, we obtain seven variants.

%Furthermore, we validate the effect of the proposed SP-loss on LIP, Pascal-Person-Part and Helen datasets, with our optimal variant setting.

\noindent\textbf{Central Clicks \emph{vs.} Random Clicks.}
All the three proposed clicking simulation strategies for training our models involve adding central clicks, and the number of such clicks corresponds to the number of human part components in a specific input image.
In this part we explore the validity of the design choice of adding a click around the central area for each human part component rather than a random click for learning to locate human parts.
Hence we build a variant trained with the ``Central-Clicks-Only'' strategy in Section~\ref{strategies}, named ``Random-Clicks-Only'', that replaces each central-type click with one randomly sampled click (``random click'') in each human body component $O^i_j$.
The trained IHP models with these two strategies are applied to IHP, and their comparison results are provided in Tab.~\ref{tab:central-vs-random}.
It is clearly shown that the model trained with ``central clicks'' significantly outperforms the model trained with ``random clicks'', with best mIoU of 68.25\% that is 12.46\% lower than that of the former.
This group of experiments well proves the effectiveness of adding the click around the central area of a human part component compared with random sampling in the first stage of training our method.

\begin{table}
    \centering
\vspace{1.5mm}
    \caption{
   \small{
Quantitative comparison between two variants trained with different initial clicking on LIP.
``Random-Clicks-Only'' denotes the model trained with ``random clicks'' while ``Central-Clicks-Only'' represents the model trained with ``central clicks''.
}}
    \resizebox{70mm}{13mm}{
    \begin{tabular}{p{2.5cm}<{\centering}|p{0.5cm}<{\centering}p{0.5cm}<{\centering}p{0.5cm}<{\centering}p{0.5cm}<{\centering}}
    \hline
Add. &0&2&4&6 \\
\hline
\hline
Random-Clicks-Only &65.78&67.46&67.78&67.90 \\
Central-Clicks-Only &79.00&80.53&80.71&80.33\\
\hline
\specialrule{0em}{3pt}{3pt}
\hline
Add. &8&10&12&15\\
\hline
\hline
Random-Clicks-Only &67.91&68.25&68.20&68.17 \\
Central-Clicks-Only &80.31&80.34&80.32&80.35 \\
\hline
    \end{tabular}
    }

\label{tab:central-vs-random}
\vspace{-5mm}
\end{table}

\noindent \textbf{Effects and Efficiency of Extra Clicks.}
We experiment on LIP, PASCAL-Person-Part and Helen datasets to explore the effects of adding extra clicks.
Among the three proposed clicking simulation strategies for training IHP models, Random-Sampling and Near-Edge-Sampling both involve the refinement stage where correction clicks are required to help the model update the initial parsing gradually till reaching our predefined parsing standards.
In this part, we first explore how the two strategies perform with a preset maximum click number, for which we experiment with three ranges including from 1 up to 15, 45, and 75. These combinations make six variant models and we also include the one trained with Central-Click-Only as a variant without any extra clicks used for training.

\begin{figure*}[t]
  \setlength{\abovecaptionskip}{-0.3cm}
\begin{center}
   \includegraphics[width=17cm,height = 4.5cm]{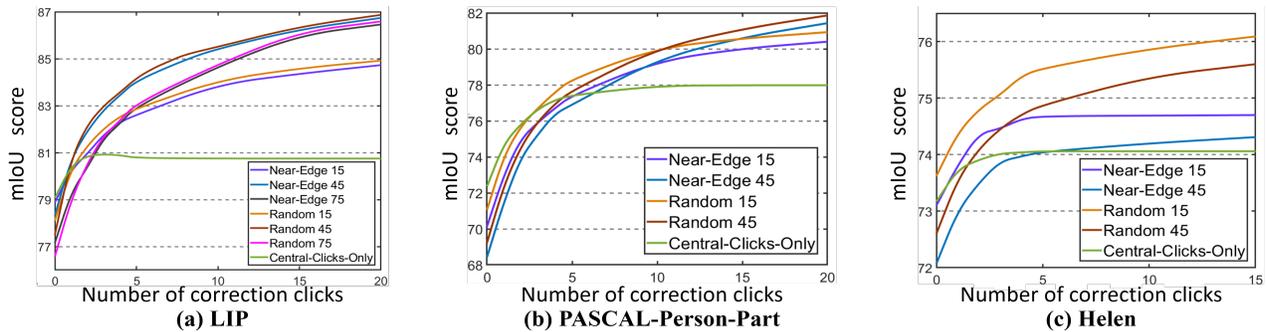}
\end{center}
\vspace{-1mm}
   \caption{
	\small{The mIoU score vs. correction clicks number on LIP, PASCAL-Person-Part and Helen respectively.
	Each curve shows the mIoU of a model varying with different correction clicks number (without SP-loss). Note that the mIoU scores at abscissa (correction clicks number) 0 represents its value of the initialization stage.
   }}
\label{fig:strategycomparisons}
\end{figure*}

\begin{figure*}[t]
  \setlength{\abovecaptionskip}{-0.2cm}
\begin{center}
   \includegraphics[width= 18cm,height = 4.6cm]{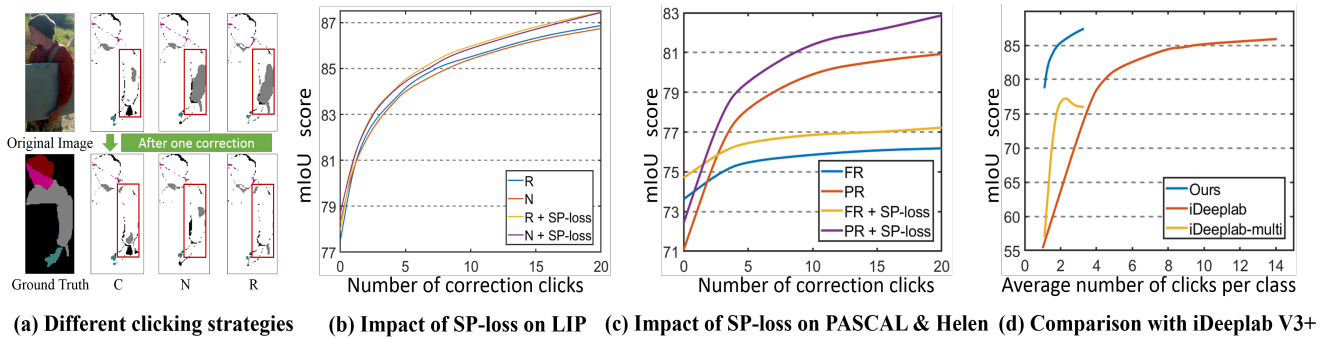}
\end{center}
\vspace{-1mm}
\caption{\small{(a) Illustration of refinement on the largest wrongly labeled region (marked by red bounding boxes) before and after adding one correction click with three clicking strategies. ``C'', ``R'', ``N'' indicate Central-Clicks-Only, Random-Sampling, Near-Edge-Sampling respectively.
The image is from LIP validation set, and all strategies are trained with $1\sim45$ EC-Num.
Different classes of wrongly labeled regions are denoted with different colors while white area indicates the correct segmentation area.
(b) Quantitative comparison of models with and without SP-loss on LIP.
``R'' and ``N'' denote performance of using Random-Sampling and Near-Edge-Sampling at $1\sim45$ EC-Num without SP-loss.
``R + SP-loss'' and  ``N + SP-loss'' denote that with SP-loss.
(c) Quantitative comparison of models with and without SP-loss on PASCAL-Person-Part and Helen.
``PR'' , ``FR'' denote performance of using Random-Sampling at $1\sim15$ EC-Num on PASCAL-Person-Part and Helen without SP-loss, while ``PR + SP-loss'' and  ``FR + SP-loss'' denote that with SP-loss.
(d) Comparison of iDeeplab V3+ with different competitors and our model on LIP.
``iDeeplab'' represents iDeeplab V3+ using its original evaluation method; ``iDeeplab-multi'' represents iDeeplab V3+ using our evaluation method; ``ours'' represents our model.
}}
\label{fig:4m1}
\end{figure*}

All the results are shown in Fig.~\ref{fig:strategycomparisons} and Tab.~\ref{tab:click-num}.
It is clearly demonstrated that with our parsing standards that are set on each dataset, the models trained with Random-Sampling outperform those with Near-Edge-Sampling given the same number of extra clicks for each group of experiments (\emph{e.g.}, with $1\sim45$ extra clicks given).
Meanwhile, the former uses less correction clicks to reach our standards, which well show its advantageous efficiency. Fig.~\ref{fig:4m1} (a) visualizes the changes of the parsing result on the largest wrongly labeled regions before and after one correction click via three clicking strategies when testing.
The shown image is picked from LIP validation set, and all strategies are trained with $1\sim45$ extra simulation clicks.
Different classes of wrongly labeled regions are denoted with different colors while white area indicates the correct segmentation area.

Specifically, when using no extra clicks for training, \emph{i.e.} the Central-Clicks-Only strategy, as shown in Fig.~\ref{fig:strategycomparisons}, the model achieves the best performance on LIP and PASCAL-Person-Part datasets, possibly due to its low interaction-dependence.
However, when dealing with face parsing on Helen dataset, it is worse than the model trained with Random-Sampling in initialization.
%This is possibly because the distribution of face parsing semantic classes also contain \textcolor{red}{other categories except for hair, and the variant from Central-Clicks-Only lacks rich simulation samples to learn the feature discrepancy between face and these categories within it.}
%face parsing的类别分布比较特殊：脸包含着除了头发以外的类别，所以当仅使用Central的时候，脸的交互点与那些在脸中的交互点缺乏丰富的training sample去学习类别间的特征差异；也就是说在训练中，face的central clicks提供的face结构信息有限，难以起到良好的定位效果。
%不太好解释
Compared with the baseline model Deeplab V3+ trained without any simulation clicks, as shown in Tab.~\ref{tab:LIP}, Tab.~\ref{tab:Pascal} and Tab.~\ref{tab:Helen},
the variant trained with Central-Clicks-Only uses only a small number of interactive clicks in initialization but achieves huge improvements on each dataset, \emph{i.e.}, 31.11\%, 10.88\%, 11.69\% on LIP, PASCAL-Person-Part and Helen, respectively.
However, the variant model based on Central-Clicks-Only cannot work well in experiments for parsing refinement with correction clicks added, as shown in Fig.~\ref{fig:strategycomparisons}, it cannot refine mIoU up to our parsing standards.
As shown in Fig.~\ref{fig:4m1} (a), the variant from Central-Clicks-Only may correct some error but cause some other in the same part.
This may be because it is only trained with central clicks and unable to tackle correction clicks.

For the Random-Sampling strategy, as shown in Fig.~\ref{fig:strategycomparisons}, with the increase of correction clicks, the mIoU of variants trained with this strategy quickly surpasses that of variants trained with Central-Clicks-Only, achieving the best performance under the same ``Add.''.
Such refinement effects mainly come from the comprehensive simulations which can benefit our model on dealing diverse correction clicks.
The variants of Random-Sampling are more dependent on correction interaction.
And their performance is not as good as the Central-Clicks-Only variants at initialization, but their excellent correction efficiency makes mIoU to a higher level.
As can be seen in Fig.~\ref{fig:4m1} (a), the Random-Sampling strategy variant can effectively correct wrong-labeled regions.
Compared with Near-Edge-Sampling, Random-Sampling can provide more diverse training samples (clicks) including large amounts of clicks within the human part and a few near-edge points, thus avoiding overfitting at boundaries.%overfitting表达正确？
~The variants of Random-Sampling can reach our standards with less correction clicks than the other two strategies.

The Near-Edge-Sampling strategy is aimed at tackling the problem that wrong-labeled regions are often around boundaries after initialization, which are hard to annotate for human users.
The variants of Near-Edge-Sampling focus more on refining the near-edge area of human parts.
As shown in Tab.~\ref{tab:nearedgeadvantage}, we compare the parsing performance within five pixels on the edge of human parts in terms of mIoU and F1-measure.
It can be seen that the performance of Near-Edge-Sampling variants is better than that of Random-Sampling ones for near-edge area parsing, demonstrating Near-Edge-Sampling is a better choice for refining parsing at human part parsing boundaries.
However, as can be seen in Fig.~\ref{fig:strategycomparisons}, the model performance of Near-Edge-Sampling is close to yet still slightly lower than that of Random-Sampling.
We guess, when fed with too many near-edge clicks during training, near-edge clicks overwhelm other types of clicks, thus dominate the model's attention and lead to a model lacking the ability to distinguish those regions relatively far away from boundaries, which are hardly recognized by using initial central clicks, as shown in Fig.~\ref{fig:4m1} (a).

To summarize, Central-Clicks-Only can do a better job in initialization, but does not work well during refinement; Near-Edge-Sampling tends to correct boundary errors while Random-Sampling contributes more to the wrongly labeled regions within human parts. With our parsing standards, the Random-Sampling strategy is a best choice. Please refer to the supplementary material for more visualization parsing results of different clicking strategies.

\begin{table}
\centering
    \vspace{1.5mm}
    \caption{
   \small{
Quantitative comparison for near-edge area between Random-Sampling and Near-Edge-Sampling on LIP dataset.
``Add.'' indicates the number of correction clicks.
``R'' indicates Random-Sampling.
``N'' indicates Near-Edge-Sampling.
}}
\resizebox{80mm}{9mm}{
\begin{tabular}{p{1cm}<{\centering}|p{0.5cm}<{\centering}p{0.5cm}<{\centering}p{0.5cm}<{\centering}p{0.5cm}<{\centering}|p{2.4cm}<{\centering}}
\hline
Add. &0&5&10&15& Evaluation method\\
\hline
\hline			
R &74.56&82.49&83.99&84.93&F1-measure \\
N &74.75&82.76&84.23&85.10&F1-measure\\
\hline
R &59.61&70.41&72.62&74.10&mIoU \\
N &60.25&71.42&73.39&74.97&mIoU \\
\hline
    \end{tabular}
    }
\label{tab:nearedgeadvantage}
\vspace{-5mm}
\end{table}
    %---------------------------------------------------------------------------------------------------------------------------------------------------

 %---------------------------------------------------------------------------------------------------------------------------------------------------
 \begin{table}[ht]
\centering
\caption{\small{
Quantitative comparison of different variants on  three benchmarks.
Best results are marked in \textbf{bold}.
``C'' denotes Central-Clicks-Only.
``N'' is Near-Edge-Sampling.
``R'' is Random-Sampling.
``SP'' is SP-loss.
``-'' indicates this model will never achieve the predefined mIoU in our parsing standards.}}
\resizebox{95mm}{15mm}{
\begin{tabular}{p{0.9cm}<{\centering}cccccc}
\hline
\specialrule{0em}{1pt}{1pt}
 & \multicolumn{2}{c}{LIP} & \multicolumn{2}{c}{PASCAL-Person-Part} & \multicolumn{2}{c}{Helen}\\
 & \multicolumn{2}{c}{(85\%mIoU 45 points)}& \multicolumn{2}{c}{(80\%mIoU 15 points)} & \multicolumn{2}{c}{(75\%mIoU 15 points)}\\
 \specialrule{0em}{1pt}{1pt}
\cline{2-3}\cline{4-5} \cline{6-7}
& Add. & Avg.& Add. & Avg. & Add. & Avg. \\
\hline
\hline
 \specialrule{0em}{1pt}{1pt}
C  & -& -& - &- &-&-\\

N  & 9 &2.19&  15& 4.46 & -&-\\

N + SP  & \textbf{7} &\textbf{1.95}& \textbf{6}&\textbf{3.02} &\textbf{1}&\textbf{1.09} \\

R  & 8& 2.07&  10 & 3.66 &3&1.27\\

R + SP& \textbf{7}& \textbf{1.95}& \textbf{6} & \textbf{3.02} &\textbf{1}&\textbf{1.09}  \\
 \specialrule{0em}{1pt}{1pt}
\hline
 \specialrule{0em}{2pt}{2pt}
\end{tabular}
}
\label{tab:click-num}
\vspace{-5mm}
\end{table}

\noindent \textbf{Effects of Different Extra Clicks Number.}
To investigate the influence of different extra clicks number (EC-Num for short) during training, here we compare the six variants by combining either Random-Sampling or Near-Edge-Sampling with different EC-Num ranges including $1\sim15$, $1\sim45$ and $1\sim75$,
under the same simulating strategy setting.
Specifically, we compare all of these variants on LIP, \emph{i.e.}, trained with $1\sim15$, $1\sim45$ and $1\sim75$ EC-Num, and four of them on PASCAL-Person-Part and Helen datasets, \emph{i.e.}, variants trained with $1\sim15$ and $1\sim45$ EC-Num. We omit the group of experiments with  $1\sim75$ on PASCAL-Person-Part and Helen since we observe the performance begins to drop at $1\sim45$ EC-Num.

From Fig.~\ref{fig:strategycomparisons}, we interestingly find that the variants provided with larger EC-Num during training (\emph{e.g.}, $1\sim75$ \emph{vs.} $1\sim45$), the performance at initialization is lower.
That means the variants trained with larger EC-Num are dependent on more interactive clicks and possess weaker recognition capacity at initialization during testing.
In addition, the performance among variants trained with different EC-Num are not always consistent.
For example, for Near-Edge-Sampling in Fig.~\ref{fig:strategycomparisons} (b), the models trained with $1\sim45$ outperform those with $1\sim15$ when the number of correction clicks is around 11.
Hence a larger EC-Num is often accompanied by poorer initialization but more efficient refinement.
A user may decide which model is more suitable according to his or her specific need.
For example, when a user requires detailed annotation (\emph{i.e.} higher mIoU), it is recommended to choose a larger EC-Num during training.
With our parsing standards, the EC-Num of $1\sim45$ is the best setting on LIP, and EC-Num of $1\sim15$ is the best setting on PASCAL-Person-Part and Helen.
We adopt these best EC-Num settings in the rest of our paper.

%---------------------------------------------------------------------------------------

\begin{table*}[ht]
\setlength{\belowcaptionskip}{-0.2cm}
\vspace{-4mm}
\centering
\caption{\small{Quantitative comparison to state-of-the-art human parsing methods and baselines on LIP validation set.
Hereinafter, ``*'' indicates we re-implement these methods with the same backbone as ours.}}
\resizebox{150mm}{45mm}{
\begin{tabular}{p{2.3cm}<{\centering}|cccccccccccccccccccccc}
\hline
Method &hat& hair& glov& sung& clot& dress& coat& sock& pant& suit& scarf\\
\hline
\hline
Attention \cite{attention} &58.87& 66.78 &23.32& 19.48& 63.20& 29.63& 49.70& 35.23& 66.04& 24.73 &12.84\\
SSL \cite{Liang2018Look} &59.75& 67.25& 28.95& 21.57 &65.30& 29.49& 51.92 &38.52& 68.02 &24.48& 14.92 \\
MuLA \cite{mula}&- &-& -& -& -& -& -& -& -& -& -\\
MMAN \cite{mman}& 57.66& 65.63& 30.07& 20.02& 64.15& 28.39& 51.98& 41.46& 71.03& 23.61& 9.65\\
PGN \cite{pgn}& 61.53& 69.13& 34.13& 26.99 &68.17& 34.93 &55.78& 42.50& 70.69 &25.30 &16.05\\
Deeplab \cite{deeplabchen}&59.46 &67.54 &32.62 &25.49& 65.78& 31.94 &55.43& 39.80 &70.45 &24.70 &15.51 \\
CE2P \cite{ce2p}&65.29 &72.54& 39.09& 32.73& 69.46& 32.52& 56.28& 49.67& 74.11& 27.23& 14.19\\
HHP \cite{wang2020hierarchical}&- &-& -& -& -& -& -& -& -& -& -\\
\hline
 \specialrule{0em}{0.5pt}{0.5pt}
  Deeplab V3+*\cite{deeplab}&63.42&69.53&33.73&29.75&67.51&33.26&54.02&44.46&71.69&25.97&18.95\\
  iDeeplab V3+*\cite{first}&81.35&76.45&61.07&55.49&84.93&69.55&84.09&75.31&82.43&76.35&75.35\\
 DEXTR* \cite{extremecut}&80.80&75.11&62.15&\textbf{74.21}&83.32&80.11&83.73&74.83&85.77&81.17&72.59\\
 Ours &\textbf{88.19}&\textbf{87.53}&\textbf{82.45}&63.91&\textbf{94.93}&\textbf{93.35}&\textbf{94.40}&\textbf{82.66}&\textbf{93.76}&\textbf{93.08}&\textbf{78.26}\\
\hline
\specialrule{0em}{3pt}{3pt}
\hline
Method & skirt& face& l-arm& r-arm& l-leg& r-leg& l-sh& r-sh& bkg& mIoU& Avg.\\
\hline
\hline
Attention \cite{attention} & 20.41& 70.58& 50.17& 54.03& 38.35& 37.70& 26.20& 27.09 &84.00 &42.92 &-\\
SSL \cite{Liang2018Look}  &24.32 &71.01& 52.64& 55.79 &40.23 &38.80 &28.08& 29.03 &84.56& 44.73 &-\\
MuLA \cite{mula}& -& -& -& -& -& -& -& -& -& 49.30 &-\\
MMAN \cite{mman}& 23.20& 69.54 &55.30& 58.13& 51.90& 52.17& 38.58& 39.05 &85.75 &46.81 &-\\
PGN \cite{pgn}&24.79& 73.74& 59.33 &60.78 &47.47 &46.62& 32.74& 33.75& 85.67 &48.51 &-\\
Deeplab \cite{deeplabchen}&28.13 &70.53& 55.76 &58.56& 48.99 &49.49& 36.76 &36.79 &85.49 &47.91 &-\\
CE2P \cite{ce2p}& 22.51& 75.50& 65.14& 66.59& 60.10& 58.59& 46.63& 46.12& 87.67& 53.10 &-\\
HHP \cite{wang2020hierarchical}& -& -& -& -& -& -& -& -& -&59.25 &-\\
\hline
 \specialrule{0em}{0.5pt}{0.5pt}
  Deeplab V3+*\cite{deeplab}&29.33&73.66&49.81&54.11&43.34&42.58&33.41&33.96&86.71&47.96 &-\\
  iDeeplab V3+*\cite{first}&86.52&81.26&77.49&78.22&82.66&78.63&71.42&71.91&90.81&77.07 &3.52\\
 DEXTR* \cite{extremecut}&89.71&86.91&83.05&83.73&87.95&86.95&\textbf{83.51}&\textbf{82.84}&\textbf{98.83}&81.86 &3.51\\
 Ours &\textbf{91.76}&\textbf{88.87}&\textbf{88.31}&\textbf{88.81}&\textbf{90.19}&\textbf{89.24}&81.56&81.43&96.28&\textbf{87.44} &3.51\\
\hline
 \specialrule{0em}{2pt}{2pt}
\end{tabular}
}
\label{tab:LIP}
\vspace{-2mm}
\end{table*}

 %-------------------------------------------------------------------------------------------------------------------------------------------------
\noindent\textbf{Effects of Semantic-Perceiving Loss.}
To prove the effectiveness of the proposed SP-loss, we compare the models with and without SP-loss on LIP, PASCAL-Person-Part and Helen datasets.
Specifically, we conduct two pairs of variants on LIP, \emph{i.e.}, Random-Sampling and Near-Edge-Sampling with or without SP-loss, and one pair on PASCAL-Person-Part and Helen, \emph{i.e.}, Random-Sampling with or without SP-loss.

From Fig.~\ref{fig:4m1} (b) and (c), we can easily find that using the proposed SP-loss brings consistent performance gains to all the variant models on the three datasets.
The improvement of SP-loss on LIP, PASCAL-Person-Part and Helen dataset is about 0.4\%-0.8\%, 1.3\%-1.9\% and 1.1\%-1.4\%, respectively. Its superiorities are relatively more obvious on PASCAL-Person-Part and Helen than those on LIP. This is possibly because the basic parsing mIoU without SP-loss on LIP are higher, and thus more difficult for models to
make improvements.
%This is possibly because the parsing standards on LIP are set higher, and thus more difficult for models to make improvements.

Furthermore, as shown in Tab.~\ref{tab:click-num}, we can see that adding the proposed SP-loss during training would save some clicks, for both ``Add.'' and ``Avg.'' of all the variant models in this group of experiments.
These improvements indicate that this loss helps the model better exploit semantics of the interactive clicks and also semantic relationships of human body parts to improve performance.

In addition, we interestingly find when the model trained without SP-loss, Near-Edge-Sampling shows slightly worse performance than Random-Sampling.
That is, SP-loss benefits more for Near-Edge-Sampling, possibly because it provides information within the human part.

%-------------------------------------------------------------------------------------------------------------------------------------------------

\begin{table*}[ht]
\vspace{-4mm}
\centering
\caption{\small{Quantitative comparison to state-of-the-art human parsing methods and baselines on PASCAL-Person-Part validation set.}}
\resizebox{130mm}{28mm}{
\begin{tabular}{p{2.5cm}<{\centering}|cccccccccc}
\hline
Method& head& torso& u-arm& l-arm& u-leg& l-leg& bkg &mIoU& Avg.	\\	
\hline
\hline
Attention \cite{attention}& 81.47& 59.06& 44.15& 42.50 &38.28& 35.62& 93.65 &56.39 &-\\
SSL \cite{Liang2018Look}& 83.26 &62.40& 47.80& 45.58& 42.32& 39.48& 94.68& 59.36 &-\\
MMAN \cite{mman}& 82.58& 62.83 &48.49& 47.37& 42.80 &40.40 &94.92& 59.91 &-\\
Structure-evolving LSTM \cite{lstm}& 82.89 &67.15& 51.42 &48.72& 51.72& 45.91 &97.18&63.57 &-\\
Deeplab-ASPP  \cite{deeplabchen}& -& -& -& -& -& -& -& 64.94 &-\\
MuLA \cite{mula}& -& -& -& -& -& -& -& 65.10 &-\\
Deeplab \cite{deeplabchen}& 85.67& 67.12& 54.00& 54.41 &47.06& 43.63& 95.16& 63.86 &-\\
PGN \cite{pgn}& \textbf{90.89} &75.12 &55.83& 64.61 &55.42& 41.57 &95.33 &68.40 &-\\
Graphonomy \cite{graphonomy}& -& -& -& -& -& -& -& 71.14 &-\\	
HHP \cite{wang2020hierarchical}& 89.73& 75.22& 66.87& 66.21& 58.69& 58.17& 96.94& 73.12 &-\\
\hline
 \specialrule{0em}{0.5pt}{0.5pt}
  Deeplab V3+*\cite{deeplab} &84.82&65.19&49.94&49.31&44.73&41.48&94.91&61.48 &-\\
  iDeeplab V3+*\cite{first} &81.48&62.06&56.36&54.41&57.89&55.05&94.04&65.90 &3.21\\
 DEXTR* \cite{extremecut}&79.97&68.32&52.29&53.21&65.74&60.44&\textbf{98.85}& 68.40 &3.34\\
 Ours&89.74&\textbf{82.16}&\textbf{74.94}&\textbf{71.09}&\textbf{75.26}&\textbf{71.29}&97.26&\textbf{80.25} &3.02	\\
\hline
 \specialrule{0em}{2pt}{2pt}
\end{tabular}
}
\label{tab:Pascal}
\end{table*}

%---------------------------------------------------------------------------------------------------------------------------------------------------

\begin{table*}[ht]
\vspace{-4mm}
\centering
\caption{\small{Quantitative comparison with baselines on Helen validation set.}}
\resizebox{170mm}{9mm}{
\begin{tabular}{p{2.3cm}<{\centering}|cccccccccccccc}
\hline
Method&face& l-brow& r-brow& l-eye& r-eye&nose&u-lip &in-mouth&l-lip&hair&bkg &mIoU&Avg.\\	
\hline
\hline
 Deeplab V3+*\cite{deeplab} &85.59&28.10&28.78&24.22&32.44&81.55&53.86&58.39&59.42&61.78&91.61&55.07 &-\\
 iDeeplab V3+*\cite{first} &74.60&33.23&39.12&61.86&60.77&80.29&14.27&24.35&36.68&65.68&91.88&52.98 &2.86\\
 Ours&\textbf{91.69}&\textbf{67.25}&\textbf{66.08}&\textbf{72.87}&\textbf{73.04}&\textbf{87.56}&\textbf{60.66}&\textbf{72.56}&\textbf{71.24}&\textbf{88.41}&\textbf{98.11}&\textbf{77.22} &2.85 \\
\hline
 \specialrule{0em}{2pt}{2pt}
\end{tabular}
}
\label{tab:Helen}
\vspace{-4mm}

\end{table*}

%-------------------------------------------------------------------------------------------------------------------------------------------------
\subsection{Comparisons to Baselines}
\label{baseline}
Since we are the first to study the IHP problem, we adopt two kinds of baselines for comparison:
1) plain ``Deeplab V3+'' \cite{deeplab} and state-of-the-art human parsing methods without interactive setting;
2) powerful standard IOS approaches including interactive object seleciton \cite{first} and DEXTR \cite{extremecut}.
We duplicate \cite{first} by replacing the backbone with Deeplab V3+, hence named ``iDeeplab V3+'' in our paper.

We compare our optimal variants (Random-Sampling+SP-loss) with the baselines on each dataset.
The results are illustrated in Tab.~\ref{tab:LIP}, Tab.~\ref{tab:Pascal} and Tab.~\ref{tab:Helen}, respectively. Please refer to supplementary material for the experimental results on CIHP.

\textbf{Comparisons to Human Parsing Methods.}
We compare our algorithm to the state-of-the-art human parsing methods to show how interactive information improves human parsing arts.
From the performance on each dataset, we can see the mIoU of our proposed method is promoted to 87.44\% in comparison with the best previous human parsing work 59.25\% on LIP, %\cite{wang2020hierarchical},
83.83\% vs. 61.05\% on CIHP and 80.25\% vs. 73.12\% on PASCAL-Person-Part.
Also, our method outperforms Deeplab V3+ by a large margin, which proves interactive information is used in our approach can further lift human parsing performance, at the cost of only a few user clicks.

\begin{figure*}[t]
  \begin{center}
   \includegraphics[width=18.5cm,height = 8cm]{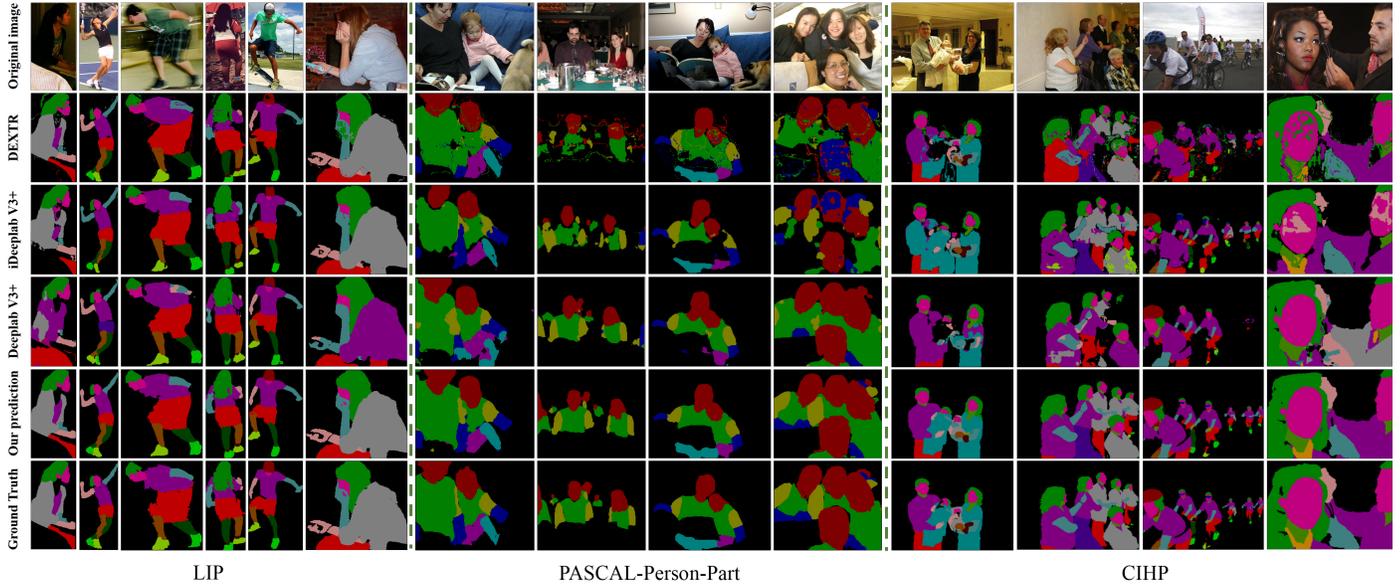}
\end{center}
\vspace{-6mm}
   \caption{\small{Illustration of parsing maps of baselines and our best variant on LIP, PASCAL-Person-Part and CIHP datasets. Note all the interactive samples are generated with the similar ``Avg.''.}}
\label{fig:long}
%\vspace{-3mm}
\end{figure*}

We observe from Tab.~\ref{tab:LIP} to Tab.~\ref{tab:Pascal} and Fig.~\ref{fig:long} that,
when no interactive information is provided, Deeplab V3+ is good at segmenting the boundaries either among adjacent human body parts or between the whole human body and background.
This is why the mIoU of background is usually the highest.
However, we also find Deeplab V3+ tends to be confused when assigning the class labels to pixels within certain human parts --- pixels in the same human part are often irregularly grouped into two or more possible categories.
An example is given in the third segmentation result of the Deeplab V3+ row in Fig.~\ref{fig:long} where the pixels in the arm part are divided into multiple groups.
We empirically observe this would happen in two cases: 1) between two similar categories, and 2) between two categories belonging to the same body part such as left and right hands, cloth and coats.
Hence, we adopt an area-based clicking mode to make up for such major defects of Deeplab V3+ by integrating explicit semantic and location information of clicks with human body parts.
Take the experimental results in Tab.~\ref{tab:LIP} for example.
Our method is most beneficial for those human parts that are difficult to distinguish but have clear contours.

%To make fair comparison, when using the same epochs as the state-of-the-art [], the performance of our model is around 2\% lower than our best performance, yet much higher than \cite{wang2020hierarchical}.
Tab.~\ref{tab:Helen} shows our method has good  generalization capacity to face parsing,
where the baseline result (Deeplab V3+) is far behind ours (55.07\% \emph{vs.} 77.22\%). Please refer to the supplementary material for visualization parsing results.%, as shown in Fig.~\ref{fig:face}.

\textbf{Comparisons to Interactive Methods.}
We compare our algorithm to two powerful standard IOS approaches, iDeeplab V3+ \cite{first} and DEXTR \cite{extremecut}.
Note iDeeplab V3+ and our method both adopt the area-based clicking mode,
while DEXTR employs an edge-based clicking mode.
For fair comparison, we align the backbones of these two baselines with our model.
To make them comparable, we test our model with around the same ``Avg.'',
\emph{i.e.}, 3.51 on LIP, 3.55 on CIHP and 3.02 on PASCAL-Person-Part.
Here our model is trained with the optimal combination of Random-Sampling, SP-loss and $1\sim45$ clicks on LIP and CIHP while  $1\sim15$ on PASCAL-Person-Part and Helen.

In evaluation, the original iDeeplab V3+ \cite{first} segments each single human part from background and then combines these binary segmentation results to obtain final mIoU, which cannot apply to our task.
Hence for fair comparison, we adopt two evaluation methods to test iDeeplab V3+.
One is the original evaluation method in \cite{first}):
1) For each current human part to segment, first an initial click is added in the central area of this part;
then we iteratively add positive and negative clicks to the central area of the greatest wrongly labeled region.
Each time a human part is segmented, we need to assign a corresponding class label to this part since traditional IOS methods only segment foreground from background without any semantics.
When all the human parts are segmented, the pixels not split into any human part are treated as background.
The other one is like below:
2) Initial clicks are required on all human parts and correction clicks are added on the greatest wrongly labeled region.
Differently from the first evaluation method, the interactive clicks in other categories can be reserved and seen as negative clicks for the current category.
The second evaluation method is similar as ours.

Seen from the Tab.~\ref{tab:LIP} to Tab.~\ref{tab:Helen},
given a close ``Avg.'',
the iDeeplab V3+ is far inferior to our method.
One of the reasons may be that interactive information of each class is not fully explored for the rest of classes in iDeeplab V3+.
The performance comparison of iDeeplab V3+ and ours on LIP is illustrated in Fig.~\ref{fig:4m1} (d).
It can be seen that the performance of iDeeplab V3+ is boosted by applying our evaluation method that well leverages interactive information among categories.
However, with the increase of interactive clicks, iDeeplab V3+ will lose the balance of positive and negative samples, and thus its performance begins to decline instead of increasing.
Meanwhile, our performance is much higher than iDeeplab V3+ when given the same ``Avg.''.%, as shown in Fig.~\ref{fig:face}.

When testing DEXTR, we assign 4 extreme points for each human part except for background,
and thus its ``Avg.'' is 3.51 on LIP, 3.59 on CIHP and 3.34 on PASCAL-Person-Part.
It has the same evaluation method with the original iDeeplab V3+.
From Tab.~\ref{tab:LIP} to Tab.~\ref{tab:Pascal} we can see that,
our method exceeds the strong baseline DEXTR by 5.58\% on LIP with the same 3.51 clicks,
7.06\% on CIHP with 3.55 vs. 3.59 clicks,
and 11.85\% on PASCAL-Person-Part with 3.02 vs. 3.34 clicks,
all achieving the best performance on most human parts,
especially for some classes difficult to recognize
such as gloves, scarf, suit on LIP /CIHP, and arm, leg on PASCAL-Person-Part.
With only 1.95 clicks shown in the last row of Tab.~\ref{tab:click-num},
our method already reaches 85\% mIoU,
which proves the efficiency of our method.
DEXTR's performance on multi-human data (CIHP and PASCAL-Person-Part) is not as good as on single-human data (LIP), indicating the edge-based clicking mode used by DEXTR is more friendly to semantic classes with continuous edges and much less sensitive to those with discrete distribution.
In contrast, our approach works well across all data sets.
%\vspace{-10mm}

The iDeeplab V3+ and DEXTR are class-agnostic, and they do not take advantage of semantic category information.
This means the two models lack the confrontation among semantic classes in training, leading to ambiguity of pixels classification along edges,
\emph{i.e.}, one pixel assigned to different body parts at the same time.
As can be seen from Fig.~\ref{fig:long}, the semantic human-part shapes they generate are not satisfactory.
In contrast, our parsing maps are more similar and even superior to ground truth.
hence our method can well assist in body-part-level manual annotations.

%-------------------------------------------------------------------------------------------------------------------------------------------------

%-------------------------------------------------------------------------------------------------------------------------------------------------

\subsection{Other Experimental Results}

\begin{figure*}[t]
    \setlength{\abovecaptionskip}{-0.5mm}
    \centering
    \includegraphics[width=18cm,height=4.8cm]{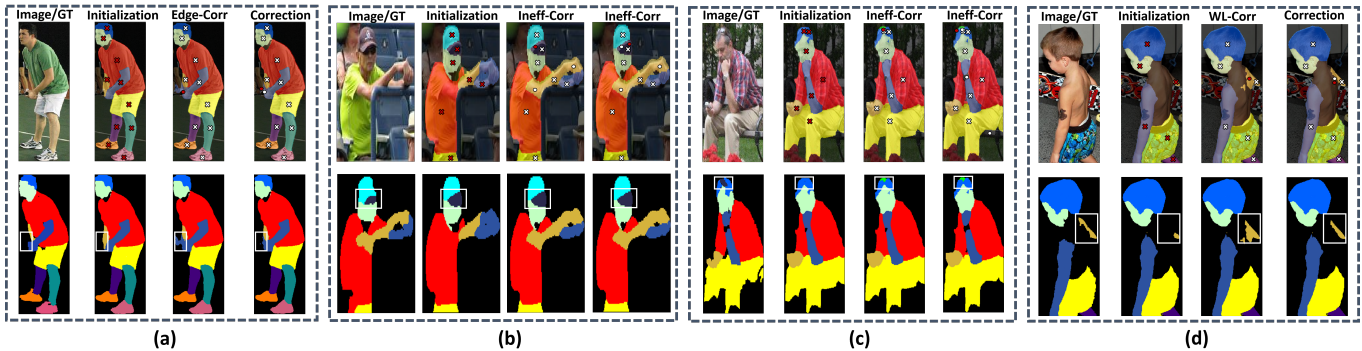}
\vspace{-1mm}
    \figcaption{
   \small{
  Real user annotation examples of LIP.
  The parsing image with user clicks are shown in the upper row. For clarity, the ground truth and parsing masks are shown in the lower row, the bounding boxes represents the noticeable mislabeled region.
``GT'' represents the ground truth of the human image.
``WL-Corr'' represents the wrongly-labeled clicks in correction phase.
  ``Edge-Corr'' represents the edge clicks in correction phase.
  ``Ineff-Corr'' represents the inefficient clicks in correction phase.
``$\times$'' denotes the clicks in initialization phase.
``$\circ$'' denotes the clicks in correction phase.
``$\times$'' and ``$\circ$'' in red indicates the new clicks at current round.
    }}
\label{fig:user-study}
\vspace{-5mm}
 \end{figure*}
\vspace{-1mm}
\noindent\textbf{User Study.}
We further conduct user study to evaluate performance of the proposed method with realistic human input.
Ten paid participants are given a tutorial on the meaning of semantic classes,
then each participant is required to label 100 images randomly but not repeatedly sampled from LIP validation set.
The labeling process includes two phase similar to our evaluation method.
1) Initialization phase:
participants need to click a point and label it for each human part component;
then our system will predict the initial parsing map.
2) Correction phase:
participants need to label one correction click around the central area of the largest wrongly labeled region for every round of correction;
then our system will feedback an updated parsing map.
The labeling process for each test image continues until 10 clicks are placed or 90\% mIoU is reached.
Over the whole labeling process, ground truth masks of the images are invisible to participants.
The model used here is ``Random-Sampling + $1 \sim 45$ points + SP-loss''.
When mIoU reaches  85\%, the ``Avg.'' is 1.86, near our simulated result 1.95.

%For each image, the average annotation time is 79.9 seconds. Specifically, in the initialization phase, the average annotation time is 33.6 seconds per-image, and one click takes about 4.7 seconds; in the correction phase, the average annotation time per-image and per-click is 44.9 and 7 seconds, respectively. As shown in Tab.~\ref{tab:user-study}, we can observe that users tend to spend more time in the correction phase, since it is more time-consuming to search the mislabeled area.

Several samples of clicks in initialization phase are illustrated in Fig.~\ref{fig:user-study}, from which we can see that users' clicking position is basically in the central area for each human part. When a human part is divided into multiple components, some users may click on each component, and some of them only click once on the prominent component, \emph{e.g.}, Fig.~\ref{fig:user-study} (c).

When conducting user study, we observe that there exist two typical kinds of failure cases caused by the following aspects: edge clicks and inefficient clicks. Failure cases caused by edge clicks are shown in the third column of Fig.~\ref{fig:user-study} (a). Concerning that pixels along the edges are naturally hard to be classified due to the semantic ambiguity, adding edge clicks for correction will cause over segmentation. However, we also observe that such failure cases can be fixed with more correction clicks, as shown in the fourth column of Fig.~\ref{fig:user-study} (a).
 Failure cases caused by inefficient clicks normally occur in regions with rare categories or narrow human components, which are depicted in the third and fourth column of Fig.~\ref{fig:user-study} (b) and Fig.~\ref{fig:user-study} (c), respectively.  The model is insufficient training on the rare categories, and the feature of narrow area is easily covered by context information because of the receptive fields of convolution kernel. These will make it difficult for the model to be efficiently guided by user clicks. We can reduce the number of this failure by improving image quality and data augmentation on rare categories.

Regardless of the aforementioned failure cases, we further explore the impact of clicking noise (\emph{i.e.}, wrongly-labeled clicks) on the parsing results. From the third column of Fig.~\ref{fig:user-study} (d) we can see that our model is sensitive to user interactions. When feeding one wrongly-labeled click, the model may produce a large mislabeled area. However, this kind of error is easier to fix by cancelling the wrongly-labeled clicks. Even if the user cannot find the wrongly-labeled clicks, the impact of these clicks can be gradually neglected by adding new correction clicks, as shown in the fourth column of Fig.~\ref{fig:user-study} (d).

Please refer to supplementary material for more details of user study.

%Here we provide the detailed annotation data about user study. Participants produce 13,578 clicks in total. During initialization phase, participants annotated 7,174 clicks, including 39 wrongly-labeled clicks. And during correction phase, participants give 6,404 clicks, including 63 wrongly-labeled clicks and 11 edge clicks. From Tab.~\ref{tab:user-study} we can conclude that the occurrence frequency of wrongly-labeled clicks or edge clicks is pretty low, and most failures can be easily corrected in the correction phase.

%------------------------------------------------------------------------------------------------------------------------------------------------------------------------
\noindent\textbf{Testing Complexity.}
The computational complexity and memory usage of our proposed model are similar to traditional human parsing models. Specifically, 1,799 MB of GPU memory is consumed during testing, and the model inference time for each image is about 20 \emph{ms} on one TITAN XP GPU. Compared with standard human parsing models without interaction, our method can maintain the similar computational complexity and memory consumption.

%------------------------------------------------------------------------------------------------------------------------------------------------------------------------
\begin{figure}[t]
    \setlength{\abovecaptionskip}{-0.5mm}
    \centering
    \includegraphics[width=6cm,height = 4cm]{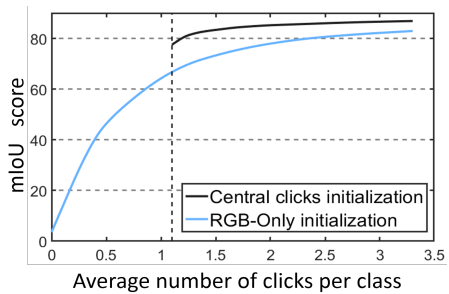}
\vspace{-1mm}
    \figcaption{
   \small{
  Comparison between models tested with or without central clicks on LIP.
    }}
\label{fig:rgbstart}
\vspace{-5mm}
 \end{figure}
\noindent\textbf{Evaluation with RGB-Only Initialization.}
Intuitively, we may consider that a more efficient way for IHP is first generate the initial mask from RGB image (\emph{i.e.}, without applying central clicks) and then require users to correct those error regions. However, based on our experiments, such a solution actually cannot guarantee both efficiency and accuracy. First, based on our user study, it usually cost a lot time to let user check the correctness of each predicted human part. Second, with consistent central clicks during both training and testing stages, the learned model will be easy to overfit human's purpose, resulting in better performance.

Fig.~\ref{fig:rgbstart} shows the quantitative comparison of these two interactive schemes using the same model (Random-Sampling+$1\sim45$). The vertical dotted line means where ``Avg.'' is the same for both interactive schemes with RGB-only (blue) and with central clicks (black) as initialization.
We can see the model only with RGB input fails to provide good initial parsing, since the model  extremely rely on interactive information and cannot only depends on RGB information to parse human images.
Its growth of mIoU is often significant at the first several correction clicks but becomes flat provided with more clicks.
Though our interactive schemes (black) relies on initial central clicks at the beginning, the final mIoU can converge at higher accuracy.

%\noindent\textbf{Failure Cases.}

\section{Conclusion}

In this paper, we propose a new challenging problem, \emph{i.e.}, IHP, making the first attempt to study human parsing under the interactive setting meanwhile further enhancing parsing performance.
Based on deep learning and inspired by conventional IOS approaches, we offer an end-to-end framework for solving the new IHP task.
We propose three clicking simulation strategies to train our IHP models, and also a SP-loss to help the model better exploit the semantics of input user clicks.
We conduct extensive experiments on four popular benchmarks and the results well prove the effectiveness of our proposed method and clicking simulation approaches for training well-performing models for tackling the IHP problem.
Considering effectiveness and parsing quality, our solution is designed to be very simple. In future, we will go beyond manually clicking on each human body part to incorporate powerful human parsing models to provide initial parsing for subsequent refinement.

\ifCLASSOPTIONcaptionsoff
  \newpage
\fi

\bibliographystyle{IEEEtran}
\bibliography{IHP}

\begin{thebibliography}{10}
\providecommand{\url}[1]{#1}
\csname url@samestyle\endcsname
\providecommand{\newblock}{\relax}
\providecommand{\bibinfo}[2]{#2}
\providecommand{\BIBentrySTDinterwordspacing}{\spaceskip=0pt\relax}
\providecommand{\BIBentryALTinterwordstretchfactor}{4}
\providecommand{\BIBentryALTinterwordspacing}{\spaceskip=\fontdimen2\font plus
\BIBentryALTinterwordstretchfactor\fontdimen3\font minus
  \fontdimen4\font\relax}
\providecommand{\BIBforeignlanguage}[2]{{%
\expandafter\ifx\csname l@#1\endcsname\relax
\typeout{** WARNING: IEEEtran.bst: No hyphenation pattern has been}%
\typeout{** loaded for the language `#1'. Using the pattern for}%
\typeout{** the default language instead.}%
\else
\language=\csname l@#1\endcsname
\fi
#2}}
\providecommand{\BIBdecl}{\relax}
\BIBdecl

\bibitem{liang2015fashion}
S.~Liu, X.~Liang, L.~Liu, K.~Lu, L.~Lin, X.~Cao, and S.~Yan, ``Fashion parsing
  with video context,'' \emph{IEEE Transactions on Multimedia}, vol.~17, no.~8,
  pp. 1347--1358, 2015.

\bibitem{liang2016clothes}
X.~Liang, L.~Lin, W.~Yang, P.~Luo, J.~Huang, and S.~Yan, ``Clothes co-parsing
  via joint image segmentation and labeling with application to clothing
  retrieval,'' \emph{IEEE Transactions on Multimedia}, vol.~18, no.~6, pp.
  1175--1186, 2016.

\bibitem{2016A}
J.~Lin, X.~Guo, J.~Shao, C.~Jiang, and S.~C. Zhu, ``A virtual reality platform
  for dynamic human-scene interaction,'' in \emph{SIGGRAPH ASIA 2016 Virtual
  Reality meets Physical Reality: Modelling and Simulating Virtual Humans and
  Environments}, 2016.

\bibitem{fang2019graspnet}
H.-S. Fang, C.~Wang, M.~Gou, and C.~Lu, ``Graspnet: A large-scale clustered and
  densely annotated datase for object grasping,'' \emph{arXiv preprint
  arXiv:1912.13470}, 2019.

\bibitem{zhang2020part}
X.~Zhang, Y.~Chen, B.~Zhu, J.~Wang, and M.~Tang, ``Part-aware context network
  for human parsing,'' \emph{In CVPR}, 2020.

\bibitem{FCM}
T.~Huang, X.~U. Yongchao, S.~Bai, Y.~Wang, and X.~Bai, ``Feature context
  learning for human parsing,'' \emph{China Science}, vol. 062, no. 012, pp.
  P.2--15, 2019.

\bibitem{Yang2018Parsing}
L.~Yang, Q.~Song, Z.~Wang, and M.~Jiang, ``Parsing r-cnn for instance-level
  human analysis.'' \emph{In CVPR}, 2018.

\bibitem{Yuan2018OCNet}
Y.~Yuan and J.~Wang, ``Ocnet: Object context network for scene parsing.''
  \emph{In arXiv preprint arXiv:1809.00916}, 2018.

\bibitem{ce2p}
T.~Ruan, T.~Liu, Z.~Huang, Y.~Wei, S.~Wei, and Y.~Zhao., ``Devil in the
  details: Towards accurate single and multiple human parsing.'' \emph{In
  AAAI}, 2019.

\bibitem{li2020self}
T.~Li, Z.~Liang, S.~Zhao, J.~Gong, and J.~Shen, ``Self-learning with
  rectification strategy for human parsing,'' \emph{In CVPR}, 2020.

\bibitem{wang2020hierarchical}
W.~Wang, H.~Zhu, J.~Dai, Y.~Pang, J.~Shen, and L.~Shao, ``Hierarchical human
  parsing with typed part-relation reasoning,'' \emph{In CVPR}, 2020.

\bibitem{graphonomy}
K.~Gong, Y.~Gao, X.~Liang, X.~Shen, M.~Wang, and L.~Lin., ``Graphonomy:
  Universal human parsing via graph transfer learning.'' \emph{In CVPR}, 2019.

\bibitem{border1}
D.~P. Papadopoulos, J.~R. Uijlings, F.~Keller, and V.~Ferrari, ``Extreme
  clicking for efficient object annotation.'' \emph{In ICCV}, 2017.

\bibitem{extremecut}
K.~K. Maninis, S.~Caelles, J.~Pont-Tuset, and L.~Van~Gool, ``Deep extreme cut:
  From extreme points to object segmentation.'' \emph{In CVPR}, 2018.

\bibitem{border2}
H.~Le, L.~Mai, B.~Price, S.~Cohen, H.~Jin, and F.~Liu, ``Interactive boundary
  prediction for object selection.'' \emph{In ECCV}, 2018.

\bibitem{border3}
D.~Acuna, H.~Ling, A.~Kar, and S.~Fidler, ``Efficient interactive annotation of
  segmentation datasets with polygon-rnn++.'' \emph{In CVPR}, 2018.

\bibitem{first}
N.~Xu, B.~Price, S.~Cohen, J.~Yang, and T.~Huang, ``Deep interactive object
  selection.'' \emph{In CVPR}, 2016.

\bibitem{region}
J.~H. Liew, Y.~Wei, W.~Xiong, S.~H. Ong, and J.~Feng, ``Regional interactive
  image segmentation networks,'' \emph{In ICCV}, 2017.

\bibitem{inside1}
Z.~Li, Q.~Chen, and V.~Koltun, ``Interactive image segmentation with latent
  diversity.'' \emph{In CVPR}, 2018.

\bibitem{full2}
E.~Agustsson, J.~R.~R. Uijlings, and V.~Ferrari, ``Interactive full image
  segmentation by considering all regions jointly.'' \emph{In CVPR}, 2019.

\bibitem{grabcut}
C.~Rother, V.~Kolmogorov, and A.~Blake, ``Grabcut: Interactive foreground
  extraction using iterated graph cut.'' \emph{In ACM TOG}, vol.~23, no.~3, pp.
  309--314, 2004.

\bibitem{superpixel}
D.~J. Chen, J.~T. Chien, H.~T. Chen, and L.~W. Chang, ``Tap and shoot
  segmentation.'' \emph{In AAAI}, 2018.

\bibitem{first-attention}
Z.~Lin, Z.~Zhang, L.-Z. Chen, M.-M. Cheng, and S.-P. Lu, ``Interactive image
  segmentation with first click attention,'' in \emph{Proceedings of the
  IEEE/CVF Conference on Computer Vision and Pattern Recognition}, 2020, pp.
  13\,339--13\,348.

\bibitem{triplet}
F.~Schroff, D.~Kalenichenko, and J.~Philbin, ``Facenet: A unified embedding for
  face recognition and clustering.'' \emph{In CVPR}, 2015.

\bibitem{Liang2018Look}
X.~Liang, K.~Gong, X.~Shen, and L.~Lin, ``Look into person:joint body parsing
  and pose estimation network and a new benchmark.'' \emph{In TPAMI}, vol.~41,
  no.~4, pp. 871--885, 2018.

\bibitem{contour1}
E.~N. Mortensen and W.~A. Barrett, ``Intelligent scissors for image
  composition.'' \emph{In Proceedings of the 22nd annual conference on Computer
  graphics and interactive techniques}, pp. 191--198, 1995.

\bibitem{contour2}
M.~Kass, A.~Witkin, and D.~Terzopoulos, ``Snakes: Active contour models.''
  \emph{In IJCV}, vol.~1, no.~4, pp. 321--331, 1988.

\bibitem{Nvidia-Object}
Z.~Wang, D.~Acuna, H.~Ling, A.~Kar, and S.~Fidler, ``Object instance annotation
  with deep extreme level set evolution,'' 2019.

\bibitem{normalizedcut}
J.~Shi and J.~Malik, ``Normalized cuts and image segmentation.'' \emph{In
  TPAMI}, vol.~22, no.~8, pp. 888--905, 2000.

\bibitem{graphcut1}
Y.~Y. Boykov and M.-P. Jolly, ``Interactive graph cuts for optimal boundary and
  region segmentation of objects in nd images.'' \emph{In ICCV}, 2001.

\bibitem{graphcut2}
V.~Vezhnevets and V.~Konouchine, ``Growcut: Interactive multi-label nd image
  segmentation by cellular automata.'' \emph{In proc. of Graphicon}, 2005.

\bibitem{graphcut3}
Y.~Li, J.~Sun, C.-K. Tang, and H.-Y. Shum, ``Lazy snapping.'' \emph{In ACM
  ToG}, vol.~23, pp. 303--308, 2004.

\bibitem{geodesics1}
X.~Bai and G.~Sapiro, ``Geodesic matting: A framework for fast interactive
  image and video segmentation and matting.'' \emph{In IJCV}, vol.~82, no.~2,
  pp. 113--132, 2009.

\bibitem{geodesics2}
A.~Criminisi, T.~Sharp, and A.~Blake, ``Geos: Geodesic image segmentation.''
  \emph{In ECCV}, 2008.

\bibitem{comb1}
B.~L. Price, B.~Morse, and S.~Cohen, ``Geodesic graph cut for interactive image
  segmentation.'' \emph{In CVPR}, 2010.

\bibitem{comb2}
V.~Gulshan, C.~Rother, A.~Criminisi, A.~Blake, and A.~Zisserman, ``Geodesic
  star convexity for interactive image segmentation.'' \emph{In CVPR}, 2010.

\bibitem{randomwalk}
L.~Grady., ``Random walks for image segmentation.'' \emph{In TPAMI}, vol.~28,
  no.~11, pp. 1768--1783, 2006.

\bibitem{FCN}
J.~Long, E.~Shelhamer, and T.~Darrell, ``Fully convolutional networks for
  semantic segmentation.'' \emph{In CVPR}, 2015.

\bibitem{gaussian1}
A.~Benard and M.~Gygli, ``Interactive video object segmentation in the wild.''
  \emph{In arXiv preprint:1801.00269}, 2017.

\bibitem{full1}
M.~Andriluka, J.~R.~R. Uijlings, and V.~Ferrari, ``Fluid annotation: A
  human-machine collaboration interface for full image annotation.'' \emph{In
  ACM Multimedia}, 2018.

\bibitem{gaussian2}
S.~Mahadevan, P.~Voigtlaender, and B.~Leibe, ``Iteratively trained interactive
  segmentation.'' \emph{In BMVC}, 2018.

\bibitem{BRS}
W.~D. Jang and C.~S. Kim, ``Interactive image segmentation via backpropagating
  refinement scheme.'' \emph{In CVPR}, 2019.

\bibitem{r-BRS}
K.~Sofiiuk, I.~Petrov, O.~Barinova, and A.~Konushin, ``f-brs: Rethinking
  backpropagating refinement for interactive segmentation,'' \emph{In CVPR},
  2020.

\bibitem{in-out-side}
S.~Zhang, J.~H. Liew, Y.~Wei, S.~Wei, and Y.~Zhao, ``Interactive object
  segmentation with inside-outside guidance,'' in \emph{Proceedings of the
  IEEE/CVF Conference on Computer Vision and Pattern Recognition}, 2020, pp.
  12\,234--12\,244.

\bibitem{deeplab}
L.~Chen, Y.~Zhu, G.~Papandreou, F.~Schroff, and H.~Adam, ``Encoder-decoder with
  atrous separable convolution for semantic image segmentation.'' \emph{In
  arXiv preprint arXiv:1802.02611}, 2018.

\bibitem{PSP}
H.~Zhao, J.~Shi, X.~Qi, X.~Wang, and J.~Jia, ``Pyramid scene parsing network.''
  \emph{In CVPR}, 2017.

\bibitem{Liang2015Human}
X.~Liang, C.~Xu, X.~Shen, J.~Yang, L.~Si, J.~Tang, L.~Lin, and S.~Yan, ``Human
  parsing with contextualized convolutional neural network.'' 2015.

\bibitem{pgn}
K.~Gong, X.~Liang, Y.~Li, Y.~Chen, M.~Yang, and L.~Lin., ``Instance-level human
  parsing via part grouping network.'' \emph{In ECCV}, 2018.

\bibitem{pascalpart}
X.~Chen, R.~Mottaghi, X.~Liu, S.~Fidler, and R.~Urtasun, ``Detect what you can:
  Detecting and representing objects using holistic models and body parts.''
  \emph{In CVPR}, 2014.

\bibitem{face}
B.~M. Smith, L.~Zhang, J.~Brandt, Z.~Lin, and J.~Yang., ``Exemplar-based face
  parsing.'' \emph{In CVPR}, 2013.

\bibitem{balancedloss1}
S.~Xie and Z.~Tu., ``Holistically-nested edge detection.'' \emph{In IJCV},
  2017.

\bibitem{balancedloss2}
K.~Maninis, J.~Pont-Tuset, P.~Arbelaez, and L.~V. Gool., ``Convolutional
  oriented boundaries: From image segmentation to high-level tasks.'' \emph{In
  TPAMI}, vol.~40, no.~4, pp. 819--833, 2017.

\bibitem{coco}
T.~Y. Lin, M.~Maire, S.~Belongie, L.~Bourdev, R.~Girshick, and J.~Hays,
  ``Microsoft coco: Common objects in context.'' \emph{In ECCV}, 2014.

\bibitem{helen}
V.~Le, J.~Brandt, Z.~Lin, L.~Bourdev, T.~Huang, P.~Lazebnik, S.and~Perona,
  Y.~Sato, and C.~Schmid, ``Interactive facial feature localization.'' \emph{In
  ECCV}, 2012.

\bibitem{imagenet}
J.~Deng, W.~Dong, R.~Socher, L.~J. Li, and F.~F. Li, ``Imagenet: a large-scale
  hierarchical image database.'' \emph{In CVPR}, 2009.

\bibitem{pytorch}
A.~Paszke, S.~Gross, S.~Chintala, G.~Chanan, E.~Yang, Z.~DeVito, Z.~Lin,
  A.~Desmaison, L.~Antiga, and A.~Lerer, ``Automatic differentiation in
  pytorch.'' 2017.

\bibitem{attention}
L.~C. Chen, Y.~Yang, J.~Wang, W.~Xu, and A.~L. Yuille, ``Attention to scale:
  Scaleaware semantic image segmentation.'' \emph{In CVPR}, 2016.

\bibitem{mula}
X.~Nie, J.~Feng, and S.~Yan, ``Mutual learning to adapt for joint human parsing
  and pose estimation.'' \emph{In ECCV}, 2018.

\bibitem{mman}
Y.~Luo, Z.~Zheng, L.~Zheng, G.~Tao, Y.~Junqing, and Y.~Yang., ``Macro-micro
  adversarial network for human parsing.'' \emph{In ECCV}, 2018.

\bibitem{deeplabchen}
L.~Chen, G.~Papandreou, I.~Kokkinos, K.~Murphy, and A.~Yuille., ``Deeplab:
  Semantic image segmentation with deep convolutional nets, atrous convolution,
  and fully connected crfs.'' \emph{In TPAMI}, vol.~40, no.~4, pp. 834--848,
  2018.

\bibitem{lstm}
X.~Liang, L.~Lin, X.~Shen, J.~Feng, S.~Yan, and E.~P. Xing., ``Interpretable
  structure-evolving lstm.'' \emph{In CVPR}, 2017.

\end{thebibliography}

\begin{IEEEbiography}[{\includegraphics[width=1.2in,height=1.2in,clip,keepaspectratio]{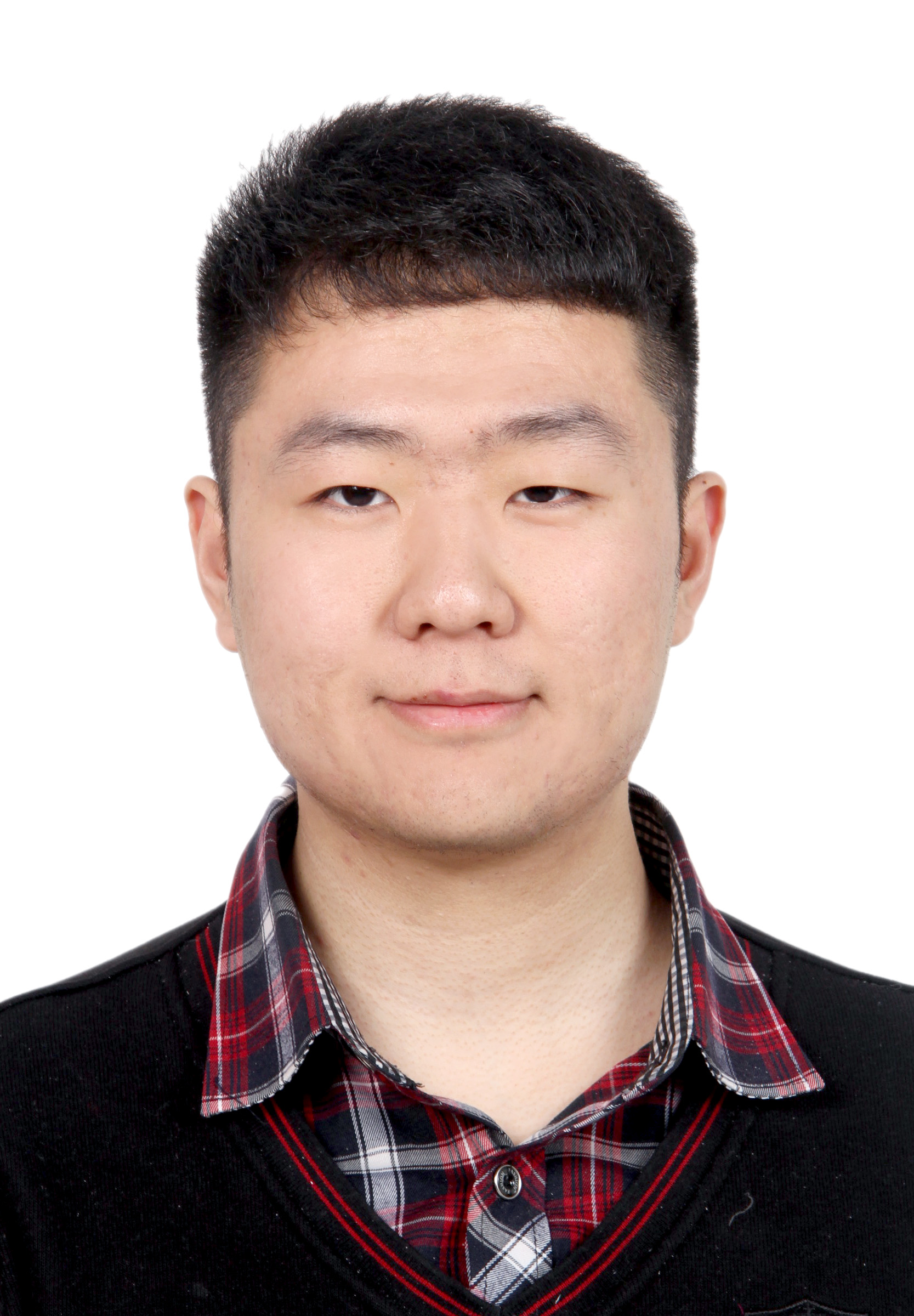}}]{Yutong~Gao} received the MA. degree from National Taipei University of Technology, Taipei, China, in 2017. received the BSc degree from Northeastern University, Liaoning, China. He is currently working toward the PhD degree in the School of Computer and Information Technology,Beijing Jiaotong University, Beijing, China.
 His current research interests include computer vision and machine learning. He is a student member of the IEEE.
\end{IEEEbiography}

\begin{IEEEbiography}[{\includegraphics[width=1.2in,height=1.2in,clip,keepaspectratio]{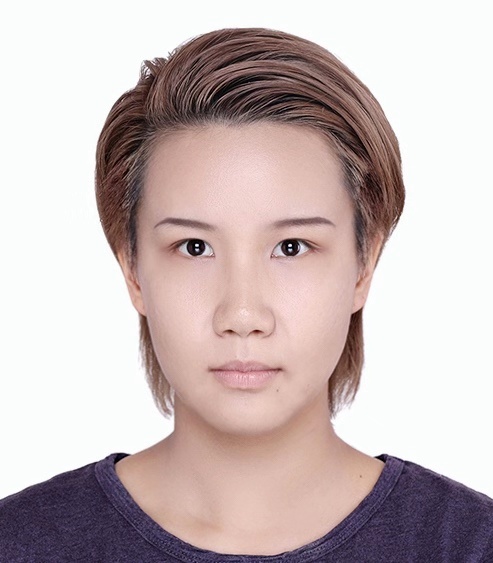}}]{Liqian~Liang}
received the BSc degree in Computer cience from the
School of Computer and Information Technology,
Beijing Jiaotong University, Beijing, China, in 2014.
Currently, she is working toward the PhD degree in the
School of Computer and Information Technology,
Beijing Jiaotong University, Beijing, China.
She has been a visiting scholar in the School of Computer Science, The University of Adelaide, Australia,
  from 2016 to 2017.
Her research interests include computer vision and machine learning.
She is a student member of the IEEE.
\end{IEEEbiography}

\begin{IEEEbiography}[{\includegraphics[width=1.2in,height=1.2in,clip,keepaspectratio]{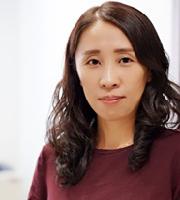}}]{Congyan~Lang}
 received the Ph.D. degree from the
School of Computer and Information Technology,
Beijing Jiaotong University, Beijing, China, in 2006.
She was a Visiting Professor with the Department
of Electrical and Computer Engineering,
National University of Singapore, Singapore,
from 2010 to 2011. From 2014 to 2015, she visited
the Department of Computer Science, University
of Rochester, Rochester, NY, USA, as a Visiting
Researcher. She is currently a Professor with
the School of Computer and Information Technology, Beijing Jiaotong
University. Her current research interests include multimedia information
retrieval and analysis, machine learning, and computer vision.
\end{IEEEbiography}

\begin{IEEEbiography}[{\includegraphics[width=1.2in,height=1.2in,clip,keepaspectratio]{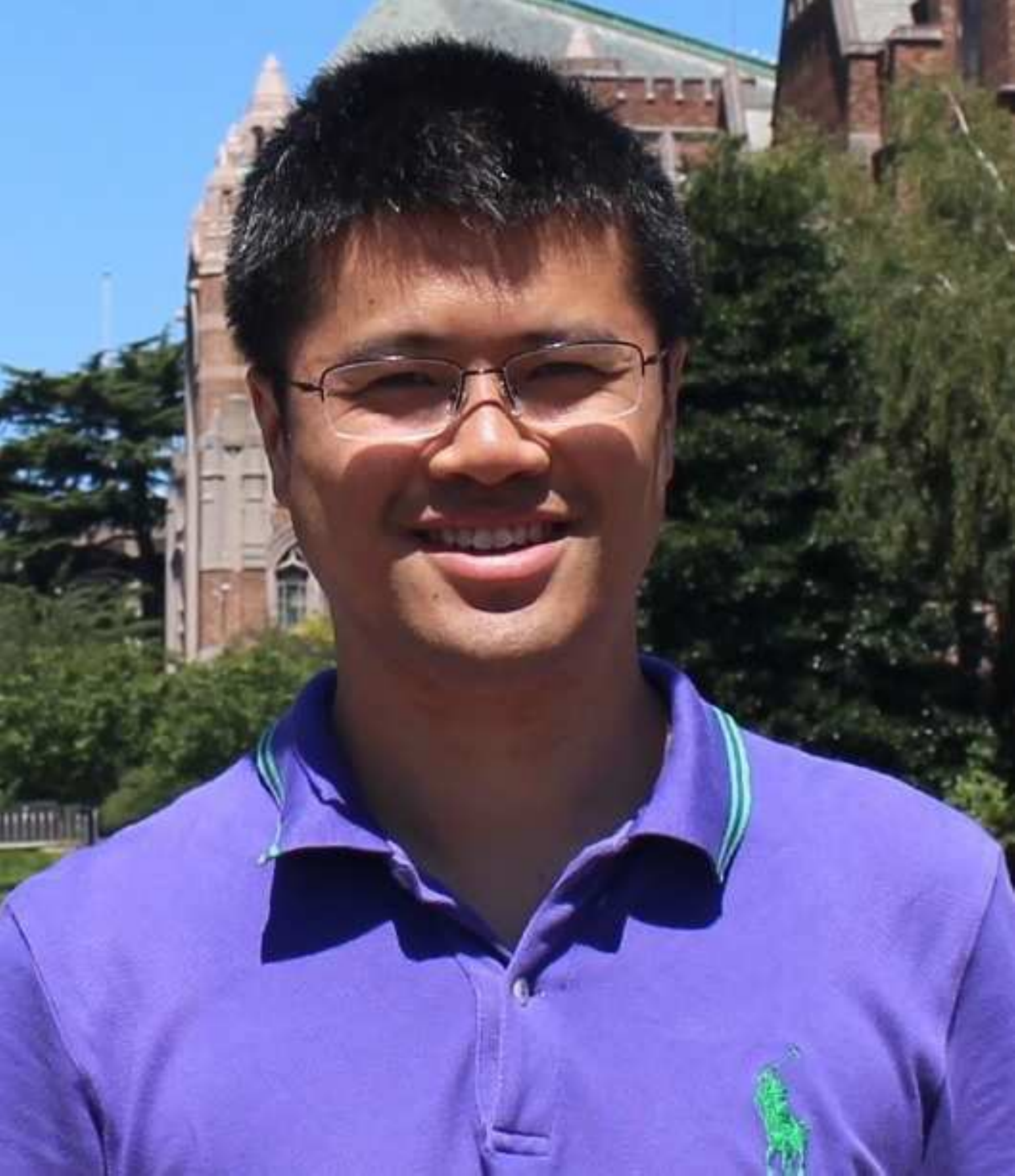}}]{Songhe~Feng}
  received the Ph.D. degree in the School of Computer and Information Technology,
  Beijing Jiaotong University, Beijing, P.R. China, in 2009.
  He is currently a Professor in the School of Computer and Information Technology,
  Beijing Jiaotong University.
  He has been a visiting scholar in the Department of Computer Science and Engineering,
  Michigan State University, USA, from 2013 to 2014.
  His research interests include computer vision and machine learning.
\end{IEEEbiography}

\begin{IEEEbiography}[{\includegraphics[width=1.2in,height=1.2in,clip,keepaspectratio]{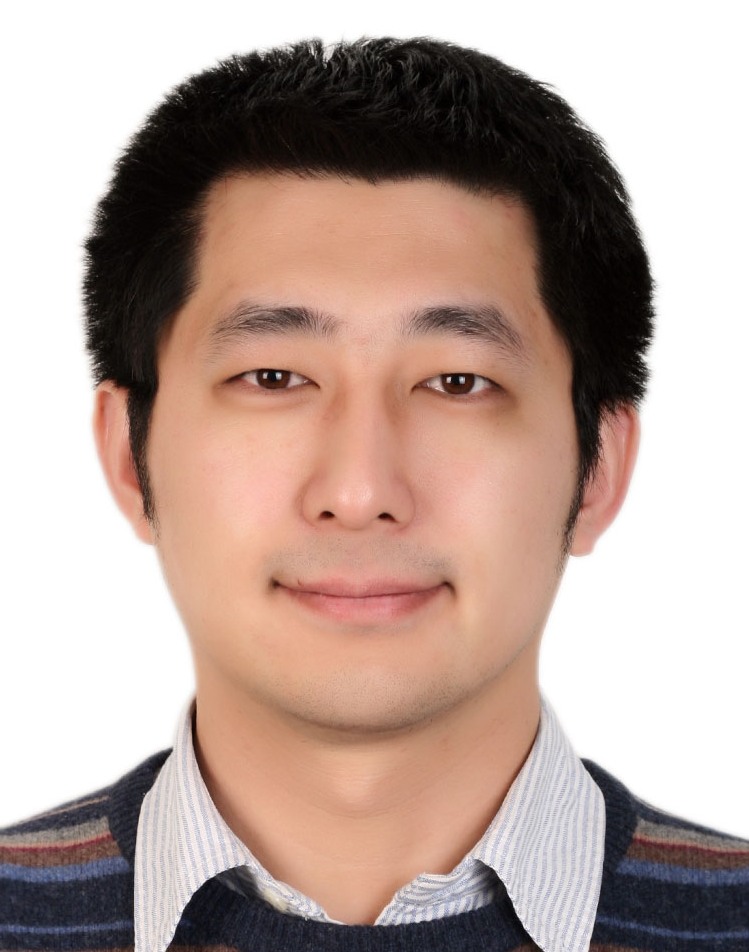}}]{Yidong Li }
is the Vice-Dean and a professor in the School of Computer and Information Technology at Beijing Jiaotong University.
Dr. Li received his B.Eng. degree in electrical and electronic engineering from Beijing Jiaotong University in 2003,
and M.Sci. and Ph.D. degrees in computer science from the University of Adelaide, in 2006 and 2010, respectively.
Dr. Li's research interests include big data analysis, privacy preserving and information security, data mining, social computing and intelligent transportation.
Dr. Li has published more than 80 research papers in various journals and refereed conferences.
He has also co-authored/co-edited 5 books (including proceedings) and contributed several book chapters.
\end{IEEEbiography}

\begin{IEEEbiography}[{\includegraphics[width=1.2in,height=1.2in,clip,keepaspectratio]{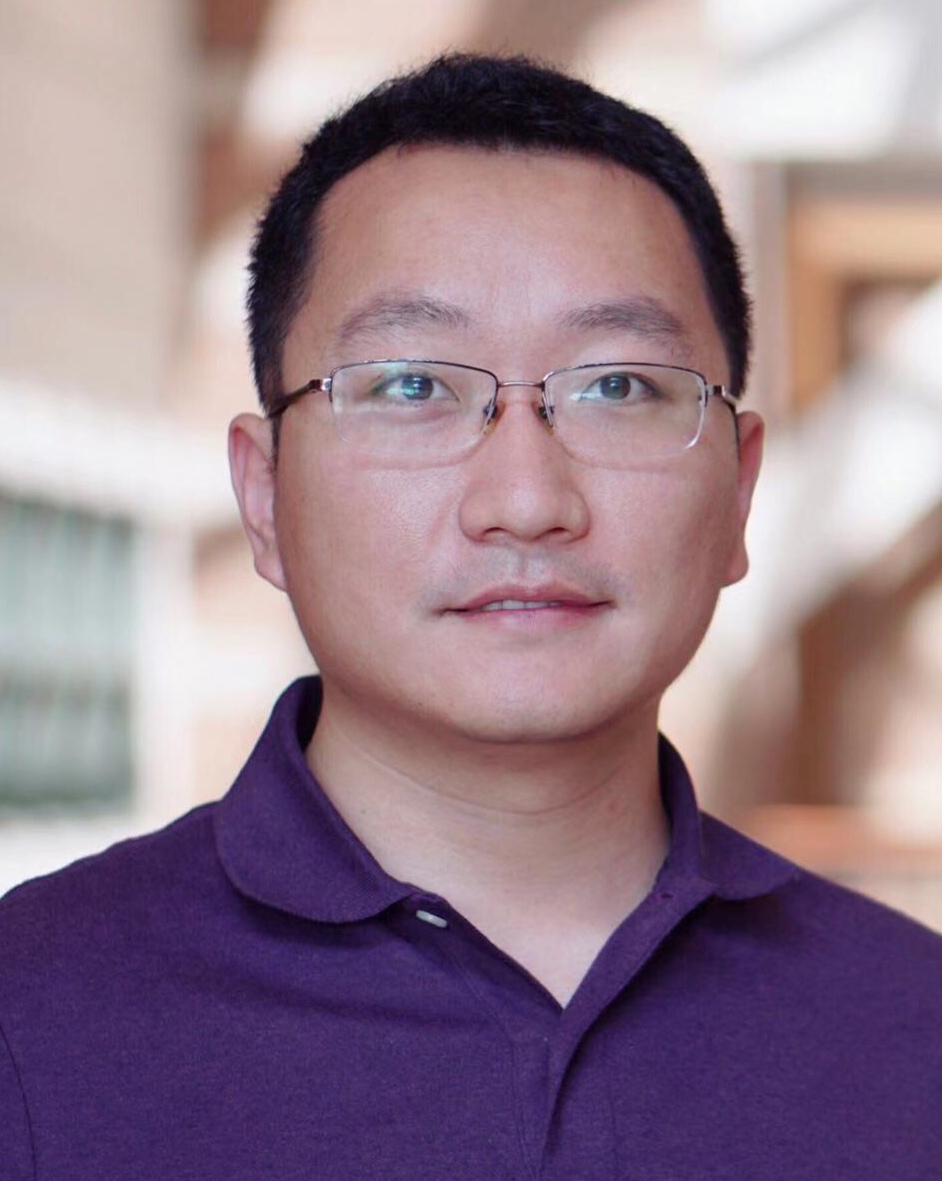}}]{Yunchao~Wei} received the Ph.D. degree from Beijing Jiaotong University, Beijing, China, in 2016. He is currently an Assistant Professor with the University of Technology Sydney, Sydney, NSW, Australia. He was a Postdoctoral Researcher with the Beckman Institute, University of Illinois at Urbana–Champaign, Urbana, IL, USA, from 2017 to 2019. His current research interests include computer vision and machine learning. Dr. Wei is an ARC Discovery Early Career Researcher Award Fellow from 2019 to 2021.
 \end{IEEEbiography}

\end{document}